\pdfoutput=1
\documentclass[10pt]{article}

\newif\ifanonymous
\anonymousfalse
\ifanonymous
\usepackage{tmlr}
\else
\usepackage[preprint]{tmlr}
\fi

\usepackage{amsmath,amssymb,amsthm}
\usepackage{microtype}
\usepackage{graphicx}
\usepackage{booktabs}
\usepackage{tabularx}
\usepackage{array}
\usepackage{xcolor}
\usepackage{tikz}
\usetikzlibrary{positioning,arrows.meta,fit,backgrounds}
\usepackage{enumitem}
\usepackage{caption}
\usepackage{subcaption}
\usepackage{url}
\usepackage[colorlinks=true,linkcolor=blue!50!black,citecolor=blue!50!black,urlcolor=blue!50!black]{hyperref}

\definecolor{cOct}{HTML}{2A78D6}
\definecolor{cGPT}{HTML}{EB6834}
\definecolor{cCla}{HTML}{1BAF7A}
\setlist{itemsep=2pt,topsep=4pt}
\newcolumntype{L}{>{\raggedright\arraybackslash}X}
\newtheorem{proposition}{Proposition}

\title{Memphis: A Case Study in Seeding an Agent Task from a Model's Real-World Attempt}
\ifanonymous
\author{}
\else
\author{\name Ibrahim A. Ahmed \email ibrahim@ibrahimahmed.ca \\
      \addr Independent researcher
      \AND
      \name Ahmed Sameh Khalil \email asameh0909@gmail.com \\
      \addr Independent researcher}
\fi

\begin{document}
\maketitle

\begin{abstract}
Agent benchmarks are usually made harder by scaling up the data their tasks are built from, yet models that do well on them still fail at real work. We propose building simulated tasks from a model's own attempt at an open-ended task in the real world, so that the model's failures in the simulation resemble its failures in reality. As a case study, we had GPT-6.1 Sol manage two human workers in Cairo over chat for five shifts, with a budget of 7,000 Egyptian pounds, to build as many USB desk lamps as it could. It built two. From the real-world attempt's transcript, we build Memphis, an evaluation in a model-simulated world. Memphis is hard: GPT-6.1 Sol delivers 1.0 lamp per trial on average and repeats its real-world mistakes, buying too few stands and giving up the search for them. With a one-page hint on where to buy parts and who can wire them, it delivers 9.2. Claude Opus 5.5 delivers 2.2 without the hint and 4.6 with it. We release the task, the trials, the audits, and a redacted record of the real-world evaluation.
\end{abstract}
\vspace{0.8em}
\begin{center}
  \begin{tikzpicture}[font=\scriptsize, txt/.style={anchor=north west, text width=1.95cm, align=left, inner sep=0pt, execute at begin node={\hyphenpenalty=10000\exhyphenpenalty=10000}}, arr/.style={-{Stealth[length=2.4mm]}, line width=1pt, black!45}]
    \draw[cOct, line width=0.8pt] (0,0.32) -- (7.95,0.32);
    \node[anchor=south west, font=\scriptsize\bfseries, text=cOct, inner sep=1pt] at (0,0.34) {REAL WORLD};
    \draw[black!55, line width=0.8pt] (8.20,0.32) -- (16.15,0.32);
    \node[anchor=south west, font=\scriptsize\bfseries, text=black!60, inner sep=1pt] at (8.20,0.34) {SIMULATION};
    \fill[cOct!7, rounded corners=4pt] (0.00,0) rectangle (3.85,-3.45);
    \node[circle, fill=cOct, text=white, inner sep=0pt, minimum size=11pt, font=\scriptsize\bfseries] at (0.33,-0.38) {1};
    \node[anchor=west, font=\footnotesize\bfseries, inner sep=0pt] at (0.58,-0.38) {The attempt};
    \fill[cOct!7, rounded corners=4pt] (4.10,0) rectangle (7.95,-3.45);
    \node[circle, fill=cOct, text=white, inner sep=0pt, minimum size=11pt, font=\scriptsize\bfseries] at (4.43,-0.38) {2};
    \node[anchor=west, font=\footnotesize\bfseries, inner sep=0pt] at (4.68,-0.38) {Transcripts};
    \fill[black!4, rounded corners=4pt] (8.20,0) rectangle (12.05,-3.45);
    \node[circle, fill=black!65, text=white, inner sep=0pt, minimum size=11pt, font=\scriptsize\bfseries] at (8.53,-0.38) {3};
    \node[anchor=west, font=\footnotesize\bfseries, inner sep=0pt] at (8.78,-0.38) {Memphis};
    \fill[black!4, rounded corners=4pt] (12.30,0) rectangle (16.15,-3.45);
    \node[circle, fill=black!65, text=white, inner sep=0pt, minimum size=11pt, font=\scriptsize\bfseries] at (12.63,-0.38) {4};
    \node[anchor=west, font=\footnotesize\bfseries, inner sep=0pt] at (12.88,-0.38) {Results};
    \begin{scope}\clip[rounded corners=3pt] (0.220,-0.72) rectangle ++(1.406,-2.5); \node[anchor=north west, inner sep=0pt] at (0.220,-0.72) {\includegraphics[height=2.5cm]{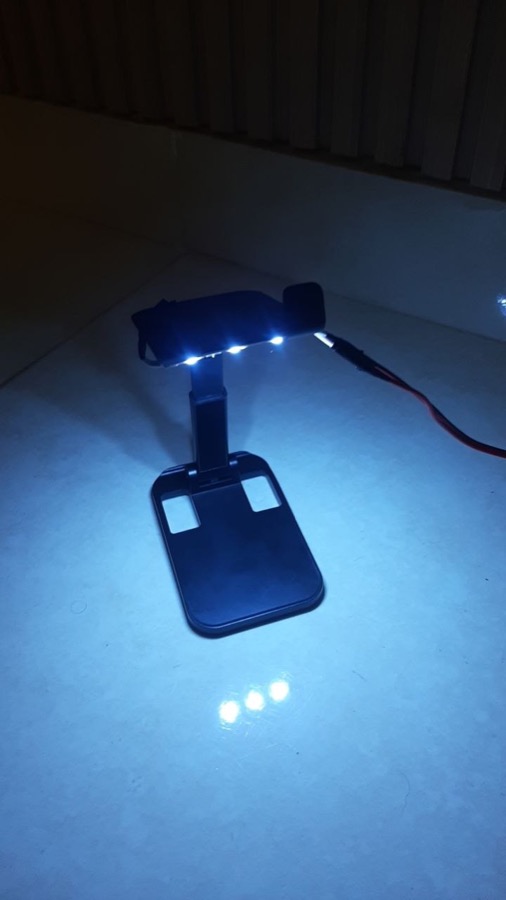}}; \end{scope}
    \node[txt] at (1.75,-0.74) {Manager: GPT-6.1 Sol\\[4pt] 2 human workers\\[4pt] 5 shifts, 7,000 EGP\\[4pt] \textbf{2 lamps}};
    \node[anchor=north west, inner sep=0pt, font=\Large\bfseries, text=cOct] at (4.40,-0.82) {798};
    \node[anchor=north west, inner sep=0pt, text width=2.1cm, execute at begin node={\hyphenpenalty=10000}] at (5.55,-0.87) {messages and photos};
    \node[anchor=north west, inner sep=0pt, font=\Large\bfseries, text=cOct] at (4.40,-1.54) {33};
    \node[anchor=north west, inner sep=0pt, text width=2.1cm, execute at begin node={\hyphenpenalty=10000}] at (5.55,-1.59) {spending requests};
    \node[anchor=north west, inner sep=0pt, font=\Large\bfseries, text=cOct] at (4.40,-2.26) {24};
    \node[anchor=north west, inner sep=0pt, text width=2.1cm, execute at begin node={\hyphenpenalty=10000}] at (5.55,-2.31) {rides};
    \node[anchor=north west, inner sep=0pt, text width=3.3cm, text=black!70] at (4.40,-2.72) {plus shops' replies and setback counts};
    \begin{scope}\clip[rounded corners=3pt] (8.420,-0.72) rectangle ++(1.406,-2.5); \node[anchor=north west, inner sep=0pt] at (8.420,-0.72) {\includegraphics[height=2.5cm]{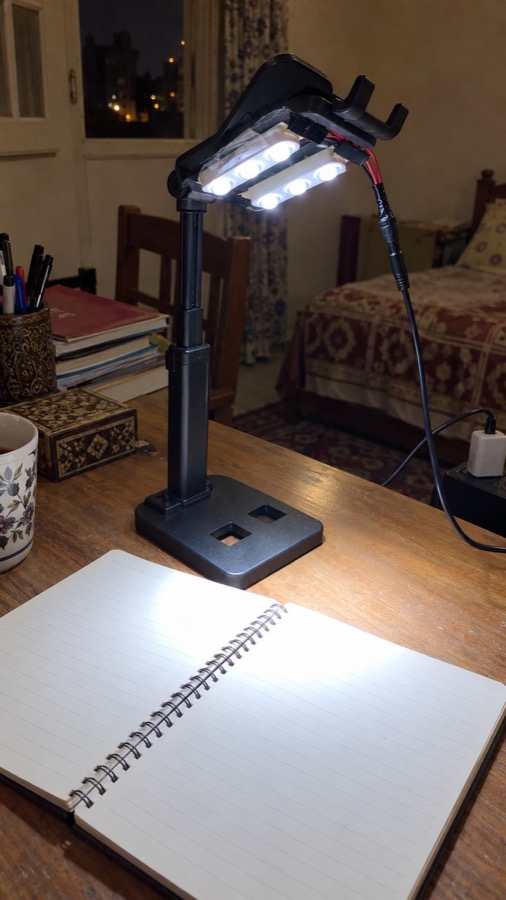}}; \end{scope}
    \node[txt] at (9.95,-0.74) {World model for workers, shops, city\\[4pt] Observed facts fixed\\[4pt] Unobserved facts sampled per seed};
    \node[anchor=north west, inner sep=0pt, text=black!70] at (12.52,-0.74) {Mean lamps per trial};
    \node[anchor=east, inner sep=0pt] at (13.17,-1.52) {Sol};
    \fill[cGPT!35] (13.25,-1.35) rectangle ++(0.252,-0.2);
    \node[anchor=west, inner sep=1pt, font=\tiny] at (13.50,-1.45) {1.0};
    \fill[cGPT] (13.25,-1.61) rectangle ++(2.321,-0.2);
    \node[anchor=west, inner sep=1pt, font=\tiny] at (15.57,-1.71) {9.2};
    \node[anchor=east, inner sep=0pt] at (13.17,-2.47) {Opus};
    \fill[cCla!35] (13.25,-2.30) rectangle ++(0.555,-0.2);
    \node[anchor=west, inner sep=1pt, font=\tiny] at (13.81,-2.40) {2.2};
    \fill[cCla] (13.25,-2.56) rectangle ++(1.161,-0.2);
    \node[anchor=west, inner sep=1pt, font=\tiny] at (14.41,-2.66) {4.6};
    \draw[black!30] (13.25,-1.25) -- (13.25,-3.05);
    \node[anchor=north west, inner sep=0pt, text width=3.4cm, text=black!70] at (12.52,-3.12) {\tikz\fill[black!25] (0,0) rectangle (7pt,4pt); no hint\quad \tikz\fill[black!70] (0,0) rectangle (7pt,4pt); hint};
    \draw[arr] (3.87,-1.72) -- (4.08,-1.72);
    \draw[arr] (7.97,-1.72) -- (8.18,-1.72);
    \draw[arr] (12.07,-1.72) -- (12.28,-1.72);
  \end{tikzpicture}
  
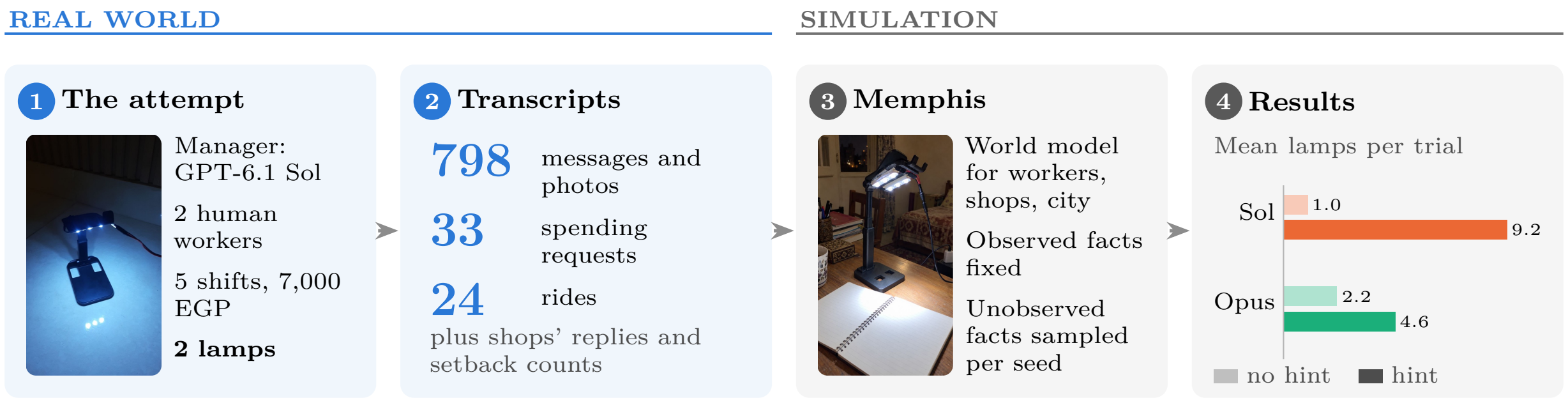
\captionof{figure}{Memphis, a simulated task built from GPT-6.1 Sol's attempt to manage two human workers. Panel 1 shows the first real lamp, and panel 3 a generated photo of the same design sent by a simulated worker. The simulated workers also send 16 images made from 14 real photos of parts, hands, packaging, and a receipt. Bars show mean lamps per trial over five seeds, with one trial per seed for Sol and three for Opus.}
  \label{fig:overview}
\end{center}

\section{Introduction}
\label{sec:intro}

Agent benchmarks increasingly test useful work.\footnote{Written with the help of Claude Opus 5.5.} SWE-bench asks models to resolve real GitHub issues \citep{jimenez2024swebench}.
GDPval grades work products from real occupations \citep{patwardhan2025gdpval}, and Vending-Bench has models run a small business over many simulated days
\citep{backlund2025vending}. Such tasks are often made harder by scaling up the seed data they are built from. SWE-Bench Pro draws enterprise-level problems from 41 business and developer-tool repositories, some of them proprietary \citep{deng2025swebenchpro}. AutomationBench spreads its tasks across several simulated business applications filled with irrelevant and sometimes misleading records \citep{shepard2026automationbench}. Longer tasks are harder still \citep{kwa2025horizons}. Despite this progress, models that do well on these benchmarks still make basic mistakes in real work, as one did when it ran a small shop in an office \citep{anthropic2025vend}. A task can be hard for reasons that do not matter in real work.

To address this, we propose seeding simulated tasks from models' own attempts at open-ended tasks in the real world. A model gets a fixed budget and a measurable objective, such as maximizing revenue, and works toward it with resources in the real world. The transcripts of this real-world evaluation are then used to build a simulated task that can be rerun. They record much of what a task author would otherwise guess, such as which shops exist, what they stock, and how long rides take. They also show where the model struggles. A simulation that reproduces the obstacles the model met should then be hard for the same reasons the real task was. The task is close to on-policy in the reinforcement-learning sense: it is built from the real attempt of the model we evaluate (Section~\ref{sec:theory}).

In October 2026, we had GPT-6.1 Sol manage two human workers in Cairo over chat for five shifts. We call the model Seb in this role. Seb had 7,000 EGP to build as many USB desk lamps from components as it could. It built two, held back more by its own decisions than by the shops or the workers (Section~\ref{sec:real}).

We built Memphis from these transcripts as a task for the Harbor agent-evaluation framework (Figure~\ref{fig:overview}). One language model, which we call the world model, simulates the workers, the shops, and the city. The simulated workers send photos from a fixed set prepared from the human workers' photos. The world model is given the chats and a file of facts about the city. The facts the workers observed are the same in every trial, and the facts they did not observe are drawn anew for each random seed. The world model can still get a fact wrong, so we check every trial (below). External setbacks are setbacks not caused by the manager's decisions, such as a shop that has sold out or a car ride that takes longer than the app estimated. The world model is told to let them happen at the rates we counted in the real-world evaluation, and we compare the simulated rates with the real ones (Section~\ref{sec:checks}). A separate model audits each trial for errors of the simulator or the grader that may have changed the number of lamps.

We evaluate two agents, GPT-6.1 Sol (hereafter Sol) and Claude Opus 5.5 (hereafter Opus), on five seeds, with and without a one-page hint of the kind a previous manager might leave. The hint tells the agent where to buy parts and who can wire them, following a plan of ours that delivers nine lamps (Appendix~\ref{app:golden}). Hinted trials show that the task can be done in the simulator, and comparing them with unhinted trials measures how much this guidance helps. We find three things. First, Sol delivers 1.0 lamp per trial on average without the hint and 9.2 with it, so much of its difficulty without the hint seems to come from not knowing where to buy parts and how to organize the work. Second, without the hint both agents often make Seb's mistakes. Sol buys too few stands and stops looking for them too soon. Opus stops looking for stands or lights while more could still be bought in 11 of its 15 trials without the hint, by our classification (Section~\ref{sec:match}). Third, Opus, which we ran on the unchanged simulator, delivers 2.2 lamps per trial without the hint and 4.6 with it (Section~\ref{sec:claude}).

We developed the simulator and the hint using runs with Sol as the agent, so both may favour Sol (Section~\ref{sec:limits}).

We also expect seeding from real-world evaluations to keep producing hard tasks as models improve, which generating environments only from task ideas that are cheap to test in a container may not. Each real-world evaluation supplies the facts for a simulated environment and shows which failures matter in it. We release a redacted record of the real-world evaluation, including the model's transcript, with the Harbor task, its code, the trials, and the audits.

\section{The real-world evaluation}
\label{sec:real}

\begin{figure}[t]
  \centering
  \begin{tikzpicture}[font=\footnotesize, day/.style={anchor=north west, text width=2.85cm, align=left, inner sep=0pt, execute at begin node={\hyphenpenalty=10000}}]
    \draw[black!25, line width=1pt] (0.06,0) -- (16.0,0);
    \filldraw[fill=white, draw=cOct, line width=1pt] (0.06,0) circle (2.4pt);
    \node[anchor=south west, inner sep=0pt] at (0.00,0.22) {\textbf{Thu 1 Oct}\hspace{0.4em}{\scriptsize\color{black!55}14--17}};
    \node[day] at (0.00,-0.28) {Seb buys 1 of the shop's 5 stands.};
    \filldraw[fill=white, draw=cOct, line width=1pt] (3.31,0) circle (2.4pt);
    \node[anchor=south west, inner sep=0pt] at (3.25,0.22) {\textbf{Sun 4 Oct}\hspace{0.4em}{\scriptsize\color{black!55}13--17}};
    \node[day] at (3.25,-0.28) {Technician next door wires the first head.};
    \fill[cOct] (6.56,0) circle (3.4pt);
    \node[anchor=south west, inner sep=0pt] at (6.50,0.22) {\textbf{Mon 5 Oct}\hspace{0.4em}{\scriptsize\color{black!55}13--17}};
    \node[day] at (6.50,-0.28) {\textbf{First lamp works.}};
    \filldraw[fill=white, draw=cOct, line width=1pt] (9.81,0) circle (2.4pt);
    \node[anchor=south west, inner sep=0pt] at (9.75,0.22) {\textbf{Tue 6 Oct}\hspace{0.4em}{\scriptsize\color{black!55}13--17}};
    \node[day] at (9.75,-0.28) {Six more heads sent across the city.};
    \fill[cOct] (13.06,0) circle (3.4pt);
    \node[anchor=south west, inner sep=0pt] at (13.00,0.22) {\textbf{Wed 7 Oct}\hspace{0.4em}{\scriptsize\color{black!55}13--17}};
    \node[day] at (13.00,-0.28) {\textbf{Second lamp}, 15:45. Seb stops at 16:07.};
  \end{tikzpicture}
  \caption{The real-world evaluation, 1 to 7 October 2026: five shifts of three to four hours. Filled dots mark the two days a lamp was finished.}
  \label{fig:timeline}
\end{figure}
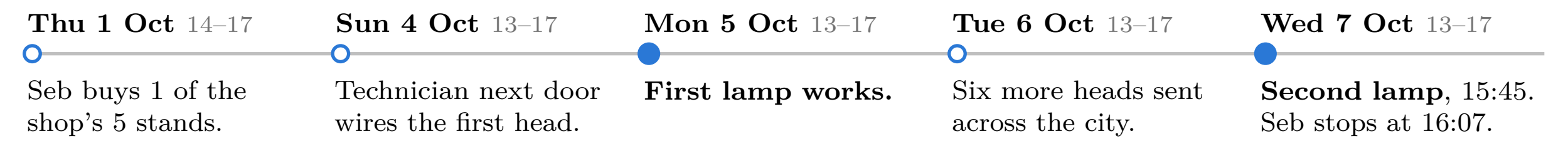

\paragraph{Setup.} Seb was GPT-6.1 Sol at maximum reasoning effort. A Telegram bot relayed
messages and photos between Seb and the two workers and posted Seb's numbered spending requests. The workers were instructed to pay only for what a posted request covered.
Seb could also email businesses, have an AI voice agent phone them, and search the web. The workers were paid for their time from a separate budget, so the 7,000 EGP covered parts, technician fees, and transport. The task asked for as many qualifying lamps as possible, delivered to a client's address by 17:00 on the last day. A qualifying lamp is a functioning, stable desk lamp that plugs into a standard 5V USB port, is
assembled during the project from components, and has a structure that holds the light in a
useful position over a desk. The workers could not be asked to cut materials or to do work that requires training, so any wiring, such as joining a 12V module to a cable, had to be done by a technician. The five shifts (Figure~\ref{fig:timeline}) came to 19 shift-hours per worker.

\paragraph{Course of the evaluation.} Seb had a working lamp by Monday afternoon and spent the remaining shifts failing to make more (Figure~\ref{fig:timeline}). The design came together quickly. After a part described over the phone turned out to be unusable, Seb sent exact part codes, and a worker bought a 12V LED module, a DC jack, and a USB step-up cable. When the shop that sold them declined to wire them, a technician next door wired them into a light head, or \emph{head}, for 50 EGP. Mounted on a 180 EGP folding phone stand, it became the first lamp. More lamps needed more stands, and Seb never got them. A worker bought six more sets of parts, which Seb had wired across the city for 600 EGP of labour and 865 EGP in rides. The only other stand was one made of card that Seb and a worker designed over four rounds of photos. It held the second and last lamp.

\paragraph{External setbacks.} We counted every external setback in the chats, meaning every setback caused by shops, rides, or the workers rather than by Seb's decisions, and kept the ids of the messages behind each count. The list is released with the task (Appendix~\ref{app:files}). Figure~\ref{fig:setbacks} shows the rates. Shops were the most common source. When a worker first asked a shop for an item, the shop lacked it, had the wrong kind, or had too few in 12 of 19 cases. Seb's remote attempts fared worse: about 60 calls and emails and four web forms produced no usable quote. Every fact Seb could act on came from the workers' own calls and visits. Our tools account for part of this. The email tool showed only the sender and subject of four suppliers' replies, and the calling tool's \$30 allowance covered only ten calls, eight of which never connected.

The two workers made about 0.45 and 0.76 mistakes per shift-hour, for example misreading an
instruction or changing route without saying so. Each spending request reserved its amount from the budget until a worker reported the payment or the cancellation, and the workers often left requests open. On the last afternoon, 17 of Seb's 33 requests were still open, Seb saw only about 100 EGP available, and 21 of the 78 messages it sent between 13:00 and 16:00 were about bookkeeping. The physical work caused little trouble. Neither of the two powered heads failed, and once a head and a stand were in the same place, a worker could assemble a lamp in 5 to 10 minutes.

\paragraph{Seb's own errors.} Seb's own decisions did more than the external setbacks to keep it at two lamps. It sent the parts for more heads to a technician 75 minutes away only a day after the first lamp worked. Besides buying
one stand out of five, it turned down five sheets of card (550 EGP) that could have been made into stands. Rides and a courier cost about two thirds of the 3,045 EGP the workers reported paying. Seb stopped responding when the model provider ran out of capacity, and it had not told the workers when to leave for the delivery address. At 16:25, a worker proposed leaving on their own.

\section{From the real-world evaluation to a simulated task}
\label{sec:task}

\subsection{Scope, tools, and score}

Memphis simulates the first two of Seb's five shifts, on Thursday and Sunday. By the end of the second shift, Seb had settled on a design but still had no lamp. A trial of these two shifts takes Sol 4 to 10 hours. The agent under evaluation takes Seb's place, with Seb's task text, its budget, and simulated versions of the two workers. It has a command shell and three tools for the task. The first sends messages, posts spending requests, and reads the clock. The second sends email and places AI phone calls. The third searches the web pages Seb saw and the shop pages generated for each seed. Everything the tools reach is simulated: the agent's container can connect only to the simulator and to the model API.

The score of a trial is the number of qualifying lamps delivered by 17:00 on Sunday, and we report the mean over the five seeds. As a reference, Appendix~\ref{app:golden} gives a plan that delivers nine lamps, and audited hinted trials reached ten or more on every seed (Appendix~\ref{app:audit}). The agent is not told either number. Its task asks for as many lamps as possible. A trial scores 0 if the agent spends more than 7,000 EGP.

\subsection{Design}
\label{sec:design}

We first wrote the simulator as code. A rule-based engine held the state of the world, and every round of audits found a new corner case, such as a worker who kept searching a district after the agent told him to stop. Each fix added a special case that sometimes broke something else. We therefore moved to a different design, in which one language model, the world model, simulates the workers, the shops, and the city, and 2,043 lines of Python handle only the few things a model does badly.

A language model, however, fills gaps differently each time it is asked. In one early trial, the world model decided that the workers could not find the delivery address without the internet, and 14 finished lamps never arrived. In another trial, the workers found it without trouble. To prevent this kind of inconsistency, we fix, before each trial begins, the facts the world model is most likely to need, and give them to it in a facts file. Anything the workers saw is copied into the file unchanged. For things they did not see but an agent may well ask about, such as other shops' stock or a technician's capacity, we draw a value from a range of plausible values. We draw it anew for each seed, so that a result that holds across seeds does not depend on a single guess about the city.

A second problem is that a language model tends to make the world too easy. In an early version, an agent without the hint reached that version's target of five lamps by taping light strips inside tubes of card, a design it had never tested that nonetheless worked every time. To bring the difficulty in line with the real-world evaluation, setbacks now occur at the rates we counted in the real-world evaluation. Worker mistakes needed a further step. When the instructions left the mistakes to the world model, it made only small ones, such as late reports (Section~\ref{sec:mistakes}). So the server, a program that sits between the agent and the world model, draws at random when a worker makes a mistake and of what kind. In its next call to the world model, it says which worker made which mistake, and the world model plays it out. The same server also keeps track of the clock and the budget.

\subsection{The facts file}

Each seed has a facts file, about 195\,KB of JSON. It does not describe the whole city. It holds what the workers saw and, drawn from ranges, what an agent is most likely to need that they did not see. The world model must apply what the file says exactly and may not invent a number, shop, or service the file does not give (Section~\ref{sec:world}). Situations the file does not cover at all are the coverage error of Section~\ref{sec:theory}. Each of the 23 businesses that the workers or Seb dealt with has an entry with the stock and prices the workers saw, what its staff said, and whom they referred the workers to. The file also gives the real sizes and shapes of the parts, where each worker lives and starts a shift, and the delivery address,
a street near a mall in Maadi. For example, the phone-accessory shop where Seb bought its stand always has five folding stands. The LED shop downtown that sold the first parts gives evasive answers before naming the technician next door.

Drawing the unobserved values anew for each seed is a form of domain randomization \citep{tobin2017domain}. Some ranges fill gaps in the 23 entries, such as prices nobody asked about and each technician's capacity and terms. Others cover the stock of nine businesses nobody dealt with, more shops in
nine districts, whether a shop answers a call, and places where the workers can sit and assemble lamps. Ride times and fares
come from a model fitted to the 24 rides the workers took.

\subsection{The world model}
\label{sec:world}

\begin{figure}[t]
  \centering
  \begin{tikzpicture}[arr/.style={-{Stealth[length=2.2mm]}, line width=0.9pt, black!50}, lab/.style={font=\scriptsize, text=black!60, inner sep=1.5pt, align=center}]
    \draw[black!55, line width=0.8pt] (0,0.32) -- (16.20,0.32);
    \node[anchor=south west, font=\scriptsize\bfseries, text=black!60, inner sep=1pt] at (0,0.34) {ONE SIMULATED TRIAL};
    \fill[black!4, rounded corners=4pt] (0.00,0.00) rectangle (4.00,-1.95);
    \node[anchor=north west, inner sep=0pt, font=\footnotesize\bfseries] at (0.22,-0.20) {Agent under test};
    \node[anchor=north west, inner sep=0pt, text width=3.60cm, align=left, font=\scriptsize, text=black!75] at (0.22,-0.62) {plays Seb, the manager\\ GPT-6.1 Sol or\\ Claude Opus 5.5\\ three task tools};
    \fill[black!4, rounded corners=4pt] (6.10,0.00) rectangle (10.10,-1.95);
    \node[anchor=north west, inner sep=0pt, font=\footnotesize\bfseries] at (6.32,-0.20) {Server (code)};
    \node[anchor=north west, inner sep=0pt, text width=3.60cm, align=left, font=\scriptsize, text=black!75] at (6.32,-0.62) {clock and money ledger\\ samples worker mistakes\\ delivers outputs on time};
    \fill[black!4, rounded corners=4pt] (12.20,0.00) rectangle (16.20,-1.95);
    \node[anchor=north west, inner sep=0pt, font=\footnotesize\bfseries] at (12.42,-0.20) {World model};
    \node[anchor=north west, inner sep=0pt, text width=3.60cm, align=left, font=\scriptsize, text=black!75] at (12.42,-0.62) {GPT-6.1 Sol\\ plays workers, shops, city\\ reads the facts file, chats\\ keeps its own notes};
    \fill[black!4, rounded corners=4pt] (0.00,-2.65) rectangle (4.00,-3.85);
    \node[anchor=north west, inner sep=0pt, font=\footnotesize\bfseries] at (0.22,-2.85) {Auditor};
    \node[anchor=north west, inner sep=0pt, text width=3.60cm, align=left, font=\scriptsize, text=black!75] at (0.22,-3.27) {separate model, reads the whole trial};
    \fill[black!4, rounded corners=4pt] (6.10,-2.65) rectangle (10.10,-3.85);
    \node[anchor=north west, inner sep=0pt, font=\footnotesize\bfseries] at (6.32,-2.85) {Grader};
    \node[anchor=north west, inner sep=0pt, text width=3.60cm, align=left, font=\scriptsize, text=black!75] at (6.32,-3.27) {separate call, reads the world model's notes};
    \fill[cOct!9, rounded corners=4pt] (12.20,-2.65) rectangle (16.20,-3.85);
    \node[anchor=north west, inner sep=0pt, font=\footnotesize\bfseries] at (12.42,-2.85) {\textcolor{cOct}{Real-world evaluation}};
    \node[anchor=north west, inner sep=0pt, text width=3.60cm, align=left, font=\scriptsize, text=black!75] at (12.42,-3.27) {798 messages, receipts, shops' answers, 24 rides};
    \draw[arr] (4.05,-0.7) -- node[lab, above] {actions} (6.05,-0.7);
    \draw[arr] (6.05,-1.25) -- node[lab, below] {messages,\\photos} (4.05,-1.25);
    \draw[arr] (10.15,-0.7) -- node[lab, above] {events,\\mistakes} (12.15,-0.7);
    \draw[arr] (12.15,-1.25) -- node[lab, below] {timed\\outputs} (10.15,-1.25);
    \draw[arr, cOct] (14.20,-2.62) -- node[lab, left, text=cOct] {facts, chats, rates} (14.20,-1.98);
    \draw[arr] (8.10,-1.98) -- node[lab, right] {Sun 17:00} (8.10,-2.62);
    \draw[arr, dashed] (6.05,-3.25) -- node[lab, above] {score} (4.05,-3.25);
  \end{tikzpicture}
  \caption{Structure of one trial. The agent and the world model never communicate directly: the server tracks time and money and delivers each of the
  world model's outputs at the time it names. The chats with the workers supply the observed facts and show the world model how these people and shops behaved.}
  \label{fig:arch}
\end{figure}
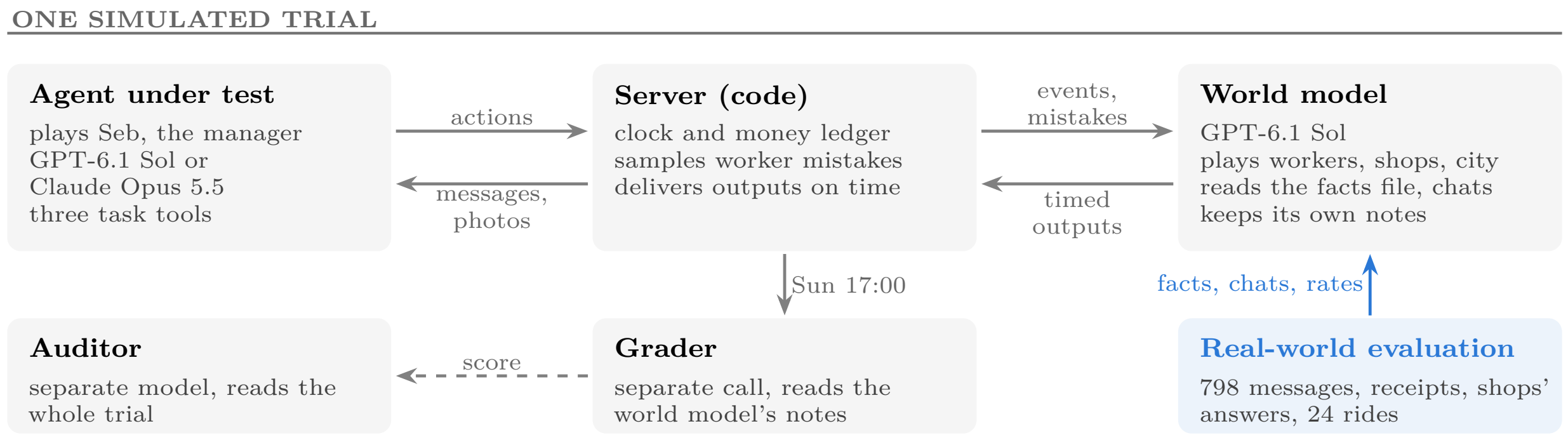

The world model is GPT-6.1 Sol at medium reasoning effort, run through Codex CLI. It simulates everything outside the agent, including the workers, shops, technicians, client, traffic, and time. Its working folder holds 65 lines of instructions, the facts file, the chats, the list of external setbacks, and its own notes on the state and history of the trial (Figure~\ref{fig:arch}). The instructions contain no rules about particular plans, shops, or outcomes.

The server also passes messages between the agent and the world model. It calls the world model whenever the agent acts, at the start and end of each shift, and at any time the model asks to be called. On each call, the model updates its notes and returns worker messages, payments, call transcripts, and emails. Each call starts a new session, so the notes are the model's only memory.
Simulated time runs at real speed while the agent works, stops while the world model reasons, and jumps ahead when the agent waits.

The instructions forbid the model to add a shop, item, price, service, or skill, and tell it not to skip a setback because the agent's plan is good or add one to punish a plan. They also keep the world physical. Each person is in one place at a time, workers know only what they have learned or can see, and parts move only when someone carries them.

\subsection{Sampled worker mistakes}
\label{sec:mistakes}

Before each call to the world model, the server decides at random whether each worker has made a mistake since the last call. It uses the rates we counted in the chats (Section~\ref{sec:real}), except that worker B's is 0.65 mistakes per shift-hour rather than the counted 0.76 (Appendix~\ref{sec:lessons}). It draws the kind of mistake in the proportions seen in the chats: misreading an instruction, taking a wrong turn, or changing route without saying so. When the world model chose the kind of mistake itself, all nine of its mistakes were late reports or detours of a few minutes, which cost less time than a typical mistake of these three kinds.
The world model then works each drawn mistake into the simulation. Trials in the final round (Section~\ref{sec:checks}) had 5 to 9 such mistakes.

\subsection{Grading and the hint}
\label{sec:hint}

At 17:00 on Sunday, a separate GPT-6.1 Sol instance acts as a judge \citep{zheng2023judging}. It reads the world model's folder, lists each unit delivered to the
address, and judges it against a written standard (Appendix~\ref{app:grading}). A
unit counts if it is one object built from components during the trial that holds a light over
an open area the size of an A4 sheet, bright enough to read and write there. It must also stand stably on the desk, be safe to handle, and arrive in time. The task tells the agent that each unit will be tested with power connected. Nothing is physically built, so
the grader decides from a unit's parts and connections whether it would light. A unit with uncertain wiring or assembly does not count.

We also run each seed with a one-page text ``left by a previous project manager'' appended to the task
(Appendix~\ref{app:hint}), which we call the hint. It gives the agent what a predecessor would know, such as where to buy parts and who can wire them, so hinted trials show that the task is achievable. We call trials with it hinted and trials without it unhinted. The hint says where to buy parts and how to make heads, and names a nearby electrician who wires them and keeps them until the workers collect them. It also says that phone stands work well but are scarce, and it lists what went wrong for earlier managers. We revised it twice with Sol as the agent. The last revision, made before the audited rounds of Section~\ref{sec:checks}, added advice about Sol's mistakes in our earlier rule-based simulator (Appendix~\ref{sec:lessons}). Because the hint was written around Sol's mistakes, it may help Sol more than other agents.

\section{Simulator error and coverage}
\label{sec:theory}

In this section, we bound how far the errors of the simulator can move an agent's score, which motivates building the task from the model's own attempt. We treat a trial as an episode of $H$ steps. Let $h_t$ be the history of everything the agent and the world have done up to step
$t$ (messages, actions, payments, and deliveries), so that no Markov assumption is needed. The agent picks its action $a_t$ from a policy
$\pi(a_t \mid h_t)$ that depends only on what it has seen. The next history follows $P(\cdot \mid h_t, a_t)$ in the real environment and $\hat P(\cdot \mid h_t, a_t)$ in Memphis, where $\hat P$ averages over the facts drawn for each seed. The score
$R(h_H) \in [0, R_{\max}]$ is read off the final history, where $R_{\max}$ is the most lamps the budget can buy. We write $J_P(\pi)$ and $J_{\hat P}(\pi)$ for the expected scores in the real environment and in Memphis, and
$\varepsilon(h,a) = \mathrm{TV}\big(P(\cdot \mid h,a),\, \hat P(\cdot \mid h,a)\big)$ for the error of the simulator on one step, with $\mathrm{TV}(p,q) = \sup_A |p(A) - q(A)|$.

\begin{proposition}
\label{prop:sim}
For any agent $\pi$,
\begin{equation}
  \big|J_P(\pi) - J_{\hat P}(\pi)\big| \;\le\; R_{\max} \sum_{t=0}^{H-1} \mathbb{E}_{P,\pi}\big[\varepsilon(h_t, a_t)\big],
  \label{eq:sim}
\end{equation}
where the expectation is over the histories and actions of the agent in the real environment.
\end{proposition}

This is a version of the simulation lemma of \citet{kearns2002near}. Appendix~\ref{app:proof} proves it by coupling the real and simulated processes. Each error enters the bound weighted by how often this agent reaches the situation in which it occurs. The simulator therefore needs to be accurate in the situations this agent reaches. Batch reinforcement learning makes a similar requirement of its data for every policy it considers, called concentrability \citep{chen2019information}. We call it coverage. We built Memphis from situations Sol reached as Seb, although Seb ran at maximum reasoning effort over five shifts and the agents here run at medium effort over two. We do not estimate the one-step errors, so the bound gives no numerical guarantee on a score.

\paragraph{Three sources of error.} Appendix~\ref{app:proof} splits the error on each step into three sources. The first is bias, how far the world model departs from the facts it was given. More real-world evaluations would not reduce it, but audits can detect departures from the facts file (Section~\ref{sec:checks}). The second is estimation error in situations the workers observed, such as a shop's stock or a rate we counted. It falls as observations accumulate. Technicians missed a promised time in 3 of 5 cases, for example, and narrowing the 90\% interval for this rate from 0.27 to 0.85 down to $\pm 0.1$ would take about 65 cases, or 13 evaluations like this one (Figure~\ref{fig:setbacks}). We check this error by comparing simulated with real setback rates. The third is coverage error, in situations the workers never observed, such as a shop drawn from our ranges. Only observing more of the city reduces it, and none of our checks measures it.

\paragraph{Coverage in the trials.} We sorted all 2,380 time-stamped entries in the world model's logs
from the ten final-round trials (Section~\ref{sec:results}) by the businesses they mention. Of the entries that name a business, 75.1\% with the hint and 67.3\% without it name only the 23 that the workers or Seb dealt with (Figure~\ref{fig:coverage}). Both percentages overstate coverage. They count mentions rather than decisions, and 59 values in the 23 entries were themselves drawn from ranges, such as how many heads the technician next door wires per day. An agent from another model
family may rely more on facts drawn from ranges, which exposes its score to more coverage error (Section~\ref{sec:claude}).

\section{Auditing the simulator and the grader}
\label{sec:checks}

We developed Memphis in rounds. In each round, we ran the current simulator on seeds 10 to 14 with and without the hint. In most rounds we also audited the trials and fixed the problems found. We numbered the rounds in order and report three audited ones: the 13th, 16th, and 18th, or r13, r16, and r18. Round r18 is the final round. We void and rerun a trial that fails for technical reasons, for example when one world-model call takes over 10 minutes (Appendix~\ref{app:audit}). No trial in the final round was voided.

\paragraph{Audits.} A separate Opus instance audited each trial. It read the input and output of every world-model call, the world model's final notes, and the agent's transcript. It looked for any point at which the simulator or the grader changed what happened, for or against the agent (Appendix~\ref{app:audit}). The agent's own decisions, and variation that a real person would also show, are not errors. Each report explains why the trial fell short of nine lamps and lists the errors found, each with the probability that it changed the count. It also gives a fair range, the lamp counts the trial could have reached without these errors, with a best estimate and an expected count.

\begin{figure}[t]
  \centering
  \begin{tikzpicture}[font=\footnotesize]
    \draw[black!12] (7.900,0.35) -- (7.900,-4.270);
    \node[below, black!60] at (7.900,-4.270) {0};
    \draw[black!12] (9.925,0.35) -- (9.925,-4.270);
    \node[below, black!60] at (9.925,-4.270) {0.25};
    \draw[black!12] (11.950,0.35) -- (11.950,-4.270);
    \node[below, black!60] at (11.950,-4.270) {0.50};
    \draw[black!12] (13.975,0.35) -- (13.975,-4.270);
    \node[below, black!60] at (13.975,-4.270) {0.75};
    \draw[black!12] (16.000,0.35) -- (16.000,-4.270);
    \node[below, black!60] at (16.000,-4.270) {1};
    \node[anchor=east, align=right, text width=7.5cm] at (7.650,0.000) {A spec claim by phone or at the counter is wrong};
    \draw[cOct, line width=1.1pt, line cap=round] (9.682,0.150) -- (13.570,0.150);
    \filldraw[fill=cOct, draw=white, line width=0.5pt] (11.500,0.150) circle (1.9pt);
    \draw[cGPT, line width=1.1pt, line cap=round] (9.277,0.000) -- (11.140,0.000);
    \filldraw[fill=cGPT, draw=white, line width=0.5pt] (10.031,0.000) circle (1.9pt);
    \draw[cCla, line width=1.1pt, line cap=round] (9.358,-0.150) -- (11.140,-0.150);
    \filldraw[fill=cCla, draw=white, line width=0.5pt] (10.110,-0.150) circle (1.9pt);
    \node[anchor=east, align=right, text width=7.5cm] at (7.650,-0.560) {The first ask for an item fails};
    \draw[cOct, line width=1.1pt, line cap=round] (11.464,-0.410) -- (14.218,-0.410);
    \filldraw[fill=cOct, draw=white, line width=0.5pt] (13.016,-0.410) circle (1.9pt);
    \draw[cGPT, line width=1.1pt, line cap=round] (14.299,-0.560) -- (14.866,-0.560);
    \filldraw[fill=cGPT, draw=white, line width=0.5pt] (14.627,-0.560) circle (1.9pt);
    \draw[cCla, line width=1.1pt, line cap=round] (14.785,-0.710) -- (15.109,-0.710);
    \filldraw[fill=cCla, draw=white, line width=0.5pt] (14.969,-0.710) circle (1.9pt);
    \node[anchor=east, align=right, text width=7.5cm] at (7.650,-1.120) {A remote call or email gives a usable quote};
    \draw[cOct, line width=1.1pt, line cap=round] (7.900,-0.970) -- (8.305,-0.970);
    \filldraw[fill=cOct, draw=white, line width=0.5pt] (7.900,-0.970) circle (1.9pt);
    \draw[cGPT, line width=1.1pt, line cap=round] (8.224,-1.120) -- (8.629,-1.120);
    \filldraw[fill=cGPT, draw=white, line width=0.5pt] (8.352,-1.120) circle (1.9pt);
    \draw[cCla, line width=1.1pt, line cap=round] (7.900,-1.270) -- (8.143,-1.270);
    \filldraw[fill=cCla, draw=white, line width=0.5pt] (7.975,-1.270) circle (1.9pt);
    \node[anchor=east, align=right, text width=7.5cm] at (7.650,-1.680) {A shop will do small custom wiring in 1--2 days};
    \draw[cOct, line width=1.1pt, line cap=round] (8.710,-1.530) -- (12.355,-1.530);
    \filldraw[fill=cOct, draw=white, line width=0.5pt] (9.925,-1.530) circle (1.9pt);
    \draw[cGPT, line width=1.1pt, line cap=round] (9.277,-1.680) -- (10.816,-1.680);
    \filldraw[fill=cGPT, draw=white, line width=0.5pt] (9.925,-1.680) circle (1.9pt);
    \draw[cCla, line width=1.1pt, line cap=round] (9.763,-1.830) -- (11.302,-1.830);
    \filldraw[fill=cCla, draw=white, line width=0.5pt] (10.478,-1.830) circle (1.9pt);
    \node[anchor=east, align=right, text width=7.5cm] at (7.650,-2.240) {A technician or shop misses a promised wiring time};
    \draw[cOct, line width=1.1pt, line cap=round] (10.087,-2.090) -- (14.785,-2.090);
    \filldraw[fill=cOct, draw=white, line width=0.5pt] (12.760,-2.090) circle (1.9pt);
    \draw[cGPT, line width=1.1pt, line cap=round] (8.953,-2.240) -- (11.059,-2.240);
    \filldraw[fill=cGPT, draw=white, line width=0.5pt] (9.769,-2.240) circle (1.9pt);
    \draw[cCla, line width=1.1pt, line cap=round] (9.277,-2.390) -- (10.978,-2.390);
    \filldraw[fill=cCla, draw=white, line width=0.5pt] (9.972,-2.390) circle (1.9pt);
    \node[anchor=east, align=right, text width=7.5cm] at (7.650,-2.800) {A technician's wiring fails};
    \draw[cOct, line width=1.1pt, line cap=round] (8.062,-2.650) -- (13.003,-2.650);
    \filldraw[fill=cOct, draw=white, line width=0.5pt] (7.900,-2.650) circle (1.9pt);
    \draw[cGPT, line width=1.1pt, line cap=round] (7.981,-2.800) -- (10.168,-2.800);
    \filldraw[fill=cGPT, draw=white, line width=0.5pt] (7.900,-2.800) circle (1.9pt);
    \draw[cCla, line width=1.1pt, line cap=round] (7.900,-2.950) -- (8.224,-2.950);
    \filldraw[fill=cCla, draw=white, line width=0.5pt] (7.900,-2.950) circle (1.9pt);
    \node[anchor=east, align=right, text width=7.5cm] at (7.650,-3.360) {A long car ride overruns the app's estimate};
    \draw[cOct, line width=1.1pt, line cap=round] (10.654,-3.210) -- (14.947,-3.210);
    \filldraw[fill=cOct, draw=white, line width=0.5pt] (13.300,-3.210) circle (1.9pt);
    \draw[cGPT, line width=1.1pt, line cap=round] (15.514,-3.360) -- (16.000,-3.360);
    \filldraw[fill=cGPT, draw=white, line width=0.5pt] (16.000,-3.360) circle (1.9pt);
    \draw[cCla, line width=1.1pt, line cap=round] (15.757,-3.510) -- (16.000,-3.510);
    \filldraw[fill=cCla, draw=white, line width=0.5pt] (15.942,-3.510) circle (1.9pt);
    \node[anchor=east, align=right, text width=7.5cm] at (7.650,-3.920) {A worker's photo needs a redo or a follow-up question};
    \draw[cOct, line width=1.1pt, line cap=round] (8.710,-3.770) -- (10.087,-3.770);
    \filldraw[fill=cOct, draw=white, line width=0.5pt] (9.275,-3.770) circle (1.9pt);
    \draw[cGPT, line width=1.1pt, line cap=round] (8.791,-3.920) -- (9.358,-3.920);
    \filldraw[fill=cGPT, draw=white, line width=0.5pt] (9.054,-3.920) circle (1.9pt);
    \draw[cCla, line width=1.1pt, line cap=round] (8.143,-4.070) -- (8.305,-4.070);
    \filldraw[fill=cCla, draw=white, line width=0.5pt] (8.216,-4.070) circle (1.9pt);
    \node[anchor=east] at (16.000,0.620) {\tikz\fill[cOct] (0,0) circle (2pt); Real world\quad \tikz\fill[cGPT] (0,0) circle (2pt); Sol, final round\quad \tikz\fill[cCla] (0,0) circle (2pt); Opus, three rounds};
    \node[black!60, anchor=north east] at (16.000,-4.55) {rate};
  \end{tikzpicture}
  \caption{Setback rates in the real-world evaluation, in Sol's final round, and in the three Opus rounds. Six of the eight rates in the final round overlap the real-world rates. Long rides overrun and first asks for an item fail more often in the simulation. Dots are observed rates and bars 90\% intervals (Beta posterior, uniform prior). Worker mistakes are omitted because the server draws them. Table~\ref{tab:setbacks} gives the counts.}
  \label{fig:setbacks}
\end{figure}
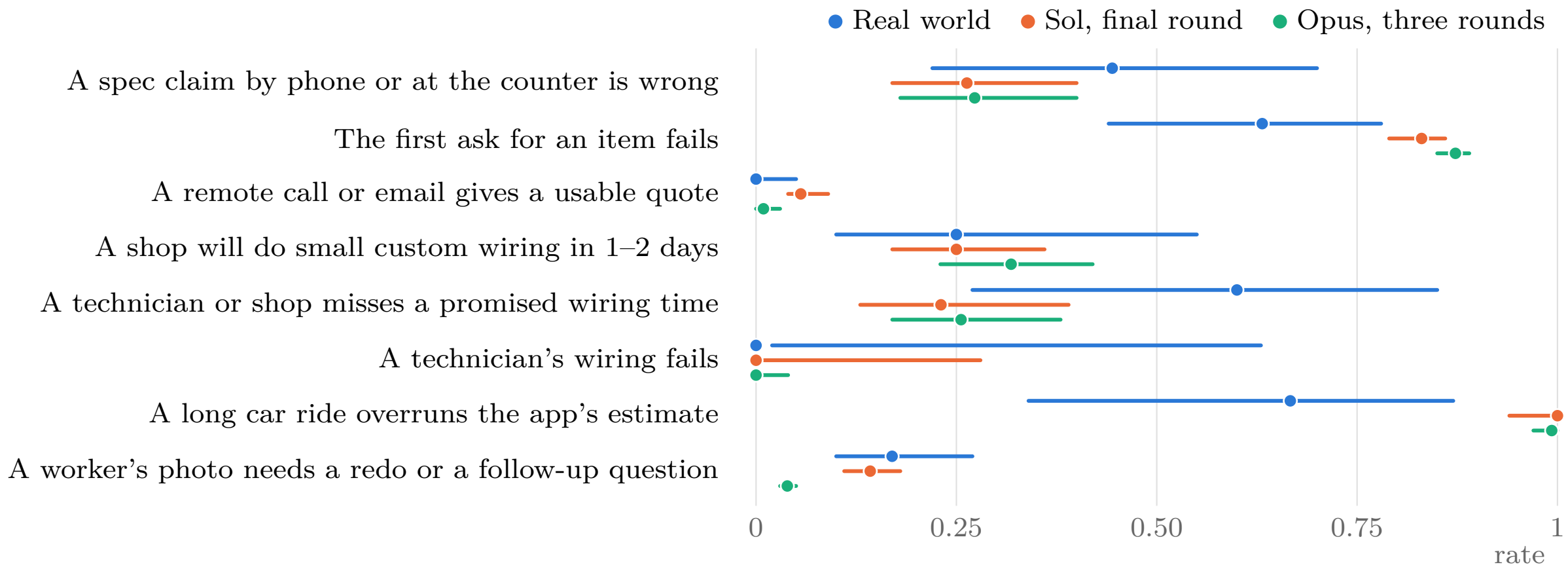

We call a trial \emph{fair} if nothing outside the agent's control changed the count, \emph{mostly fair} if the auditor names one possible error but judges it unlikely to have changed the count, \emph{partly fair} if an error plausibly changed the count, and \emph{not fair} if one probably did. Fair means faithful to the facts file and, where the facts leave the outcome open, to the setback rates of Figure~\ref{fig:setbacks}.

\paragraph{When we stopped changing the simulator.} Before the final round, we set three conditions for deciding when to stop changing the simulator, and we stopped at the first round whose ten trials met them. First, no audit judges that an error probably changed a lamp count. Errors that plausibly changed one are allowed and reported with their probability. Second, each trial's count lies within one lamp of its fair range. Third, the auditors' expected counts are on average within half a lamp per trial of the actual counts, for the hinted and the unhinted trials separately. When a round failed, we fixed the problem with a rule that applies to every trial and ran a new round. We tightened the conditions during development, and round r18 was the first to meet them. We stopped there and call that version the final simulator (tagged v12.0-rc17 in the release). Opus ran on it unchanged. The Limitations list errors we found after we stopped.

\paragraph{Setback rates.} The audits cannot tell whether the facts and rates themselves match the real-world evaluation. To check them, separate Opus instances went through the ten final-round trials and listed each occasion on which a setback of each kind could have occurred, and whether it did. They used the definitions we applied to the chats, and Appendix~\ref{app:files} gives their instructions and the classified trials. We checked 14 of their decisions by hand, covering all eight kinds, and agreed with 13. In the other, a shop agreed on the phone to prepare lamp heads but would not give a price or a time, and the instances counted this as a usable answer where we did not. Figure~\ref{fig:setbacks} names the eight kinds. For six of them, the 90\% intervals of the simulated and real rates overlap.

Two kinds of setback happened more often in the simulation. All 44 long rides in the simulation, those quoted at 25 minutes or more, ran late, against 4 of 6 in the chats, although the median ride took 1.57 times its quoted time, within the real range of 1.5 to 2. A worker's first request for an item also failed more often than in the chats, most of all for stands.

The rate of missed wiring times revealed an error in the facts file. Every technician, including those drawn from ranges, should be late as often as the real ones. In the facts file, however, only the technician across the city who wired Seb's batch is ever late. The simulated technicians who did the wiring kept all 20 of their promised times. All six missed times came from the LED shop downtown, which withdrew its offer to wire heads within an hour. Each simulated technician did what its entry said, so no audit
could flag the error. The error made plans that rely on
the technician next door easier than they were for Seb, and we estimate its effect in the Limitations.

\section{Experiments}
\label{sec:results}

\begin{figure}[t]
  \centering
  \begin{tikzpicture}[font=\footnotesize]
    \node[anchor=south west, font=\footnotesize\bfseries] at (0.000,3.660) {With the hint};
    \draw[black!12] (0.000,0.000) -- (7.300,0.000);
    \node[anchor=east, black!60] at (-0.050,0.000) {0};
    \draw[black!12] (0.000,0.810) -- (7.300,0.810);
    \node[anchor=east, black!60] at (-0.050,0.810) {3};
    \draw[black!12] (0.000,1.620) -- (7.300,1.620);
    \node[anchor=east, black!60] at (-0.050,1.620) {6};
    \draw[black!12] (0.000,2.430) -- (7.300,2.430);
    \node[anchor=east, black!60] at (-0.050,2.430) {9};
    \draw[black!12] (0.000,3.240) -- (7.300,3.240);
    \node[anchor=east, black!60] at (-0.050,3.240) {12};
    \draw[black!80, dashed, line width=0.7pt] (0.000,2.430) -- (7.300,2.430);
    \node[black!60] at (0.730,-0.3) {seed 10};
    \filldraw[fill=cGPT, draw=white, line width=0.4pt] (0.370,2.700) circle (2.1pt);
    \filldraw[fill=cCla, draw=white, line width=0.4pt] (0.650,0.000) circle (2.1pt);
    \filldraw[fill=cCla, draw=white, line width=0.4pt] (0.850,0.000) circle (2.1pt);
    \filldraw[fill=cCla, draw=white, line width=0.4pt] (1.050,0.000) circle (2.1pt);
    \node[black!60] at (2.190,-0.3) {seed 11};
    \draw[cGPT, line width=0.9pt, fill=white] (1.830,1.955) -- ++(2.3pt,-3.6pt) -- ++(-4.6pt,0) -- cycle;
    \draw[cCla, line width=0.9pt, fill=white] (2.110,0.000) circle (1.8pt);
    \filldraw[fill=cCla, draw=white, line width=0.4pt] (2.310,1.890) circle (2.1pt);
    \filldraw[fill=cCla, draw=white, line width=0.4pt] (2.510,1.890) circle (2.1pt);
    \node[black!60] at (3.650,-0.3) {seed 12};
    \filldraw[fill=cGPT, draw=white, line width=0.4pt] (3.290,2.430) circle (2.1pt);
    \draw[cCla, line width=0.9pt, fill=white] (3.570,2.225) -- ++(2.3pt,-3.6pt) -- ++(-4.6pt,0) -- cycle;
    \filldraw[fill=cCla, draw=white, line width=0.4pt] (3.770,1.620) circle (2.1pt);
    \filldraw[fill=cCla, draw=white, line width=0.4pt] (3.970,2.430) circle (2.1pt);
    \node[black!60] at (5.110,-0.3) {seed 13};
    \draw[cGPT, line width=0.9pt, fill=white] (4.750,3.510) circle (1.8pt);
    \filldraw[fill=cCla, draw=white, line width=0.4pt] (5.030,0.000) circle (2.1pt);
    \filldraw[fill=cCla, draw=white, line width=0.4pt] (5.230,3.240) circle (2.1pt);
    \filldraw[fill=cCla, draw=white, line width=0.4pt] (5.430,3.240) circle (2.1pt);
    \node[black!60] at (6.570,-0.3) {seed 14};
    \filldraw[fill=cGPT, draw=white, line width=0.4pt] (6.210,1.890) circle (2.1pt);
    \draw[cCla, line width=0.9pt, fill=white] (6.490,0.000) circle (1.8pt);
    \filldraw[fill=cCla, draw=white, line width=0.4pt] (6.690,1.350) circle (2.1pt);
    \filldraw[fill=cCla, draw=white, line width=0.4pt] (6.890,0.810) circle (2.1pt);
    \node[anchor=south west, font=\footnotesize\bfseries] at (8.000,3.660) {Without the hint};
    \draw[black!12] (8.000,0.000) -- (15.300,0.000);
    \draw[black!12] (8.000,0.810) -- (15.300,0.810);
    \draw[black!12] (8.000,1.620) -- (15.300,1.620);
    \draw[black!12] (8.000,2.430) -- (15.300,2.430);
    \draw[black!12] (8.000,3.240) -- (15.300,3.240);
    \draw[black!80, dashed, line width=0.7pt] (8.000,2.430) -- (15.300,2.430);
    \node[black!60] at (8.730,-0.3) {seed 10};
    \filldraw[fill=cGPT, draw=white, line width=0.4pt] (8.370,1.080) circle (2.1pt);
    \filldraw[fill=cCla, draw=white, line width=0.4pt] (8.650,0.000) circle (2.1pt);
    \draw[cCla, line width=0.9pt, fill=white] (8.850,0.000) circle (1.8pt);
    \filldraw[fill=cCla, draw=white, line width=0.4pt] (9.050,0.540) circle (2.1pt);
    \node[black!60] at (10.190,-0.3) {seed 11};
    \filldraw[fill=cGPT, draw=white, line width=0.4pt] (9.830,0.000) circle (2.1pt);
    \filldraw[fill=cCla, draw=white, line width=0.4pt] (10.110,0.000) circle (2.1pt);
    \filldraw[fill=cCla, draw=white, line width=0.4pt] (10.310,0.540) circle (2.1pt);
    \filldraw[fill=cCla, draw=white, line width=0.4pt] (10.510,0.810) circle (2.1pt);
    \node[black!60] at (11.650,-0.3) {seed 12};
    \filldraw[fill=cGPT, draw=white, line width=0.4pt] (11.290,0.000) circle (2.1pt);
    \filldraw[fill=cCla, draw=white, line width=0.4pt] (11.570,0.000) circle (2.1pt);
    \draw[cCla, line width=0.9pt, fill=white] (11.770,0.605) -- ++(2.3pt,-3.6pt) -- ++(-4.6pt,0) -- cycle;
    \filldraw[fill=cCla, draw=white, line width=0.4pt] (11.970,2.160) circle (2.1pt);
    \node[black!60] at (13.110,-0.3) {seed 13};
    \draw[cGPT, line width=0.9pt, fill=white] (12.750,0.270) circle (1.8pt);
    \filldraw[fill=cCla, draw=white, line width=0.4pt] (13.030,2.160) circle (2.1pt);
    \filldraw[fill=cCla, draw=white, line width=0.4pt] (13.230,0.000) circle (2.1pt);
    \filldraw[fill=cCla, draw=white, line width=0.4pt] (13.430,0.000) circle (2.1pt);
    \node[black!60] at (14.570,-0.3) {seed 14};
    \filldraw[fill=cGPT, draw=white, line width=0.4pt] (14.210,0.000) circle (2.1pt);
    \draw[cCla, line width=0.9pt, fill=white] (14.490,0.000) circle (1.8pt);
    \filldraw[fill=cCla, draw=white, line width=0.4pt] (14.690,0.810) circle (2.1pt);
    \filldraw[fill=cCla, draw=white, line width=0.4pt] (14.890,1.350) circle (2.1pt);
    \node[anchor=east, black!60, rotate=90, anchor=south] at (-0.45,1.620) {lamps};
    \node[anchor=east, black!70] at (15.300,2.600) {target};
    \node[anchor=north] at (7.650,-0.6) {\tikz\filldraw[fill=cGPT,draw=white] (0,0) circle (2.1pt); Sol, final round\quad \tikz\filldraw[fill=cCla,draw=white] (0,0) circle (2.1pt); Opus, rounds c1 to c3\qquad \tikz\fill[black!70] (0,0) circle (2.1pt); fair\quad \tikz\draw[black!70, line width=0.9pt] (0,0) circle (1.8pt); mostly fair\quad \tikz\draw[black!70, line width=0.9pt] (0,0) -- ++(2.3pt,-3.6pt) -- ++(-4.6pt,0) -- cycle; partly fair or not fair};
  \end{tikzpicture}
  \caption{Without the hint, both agents deliver far fewer than nine lamps (dashed) in most trials. With it, Sol reaches nine lamps on three seeds, and Opus's three trials on a seed often differ widely. Lamps delivered per trial on the final simulator: Sol once per seed, Opus three times (rounds c1 to c3, left to right). Marker shape is the audit verdict (Section~\ref{sec:checks}). Tables~\ref{tab:main} and~\ref{tab:claude} give the counts.}
  \label{fig:results}
\end{figure}
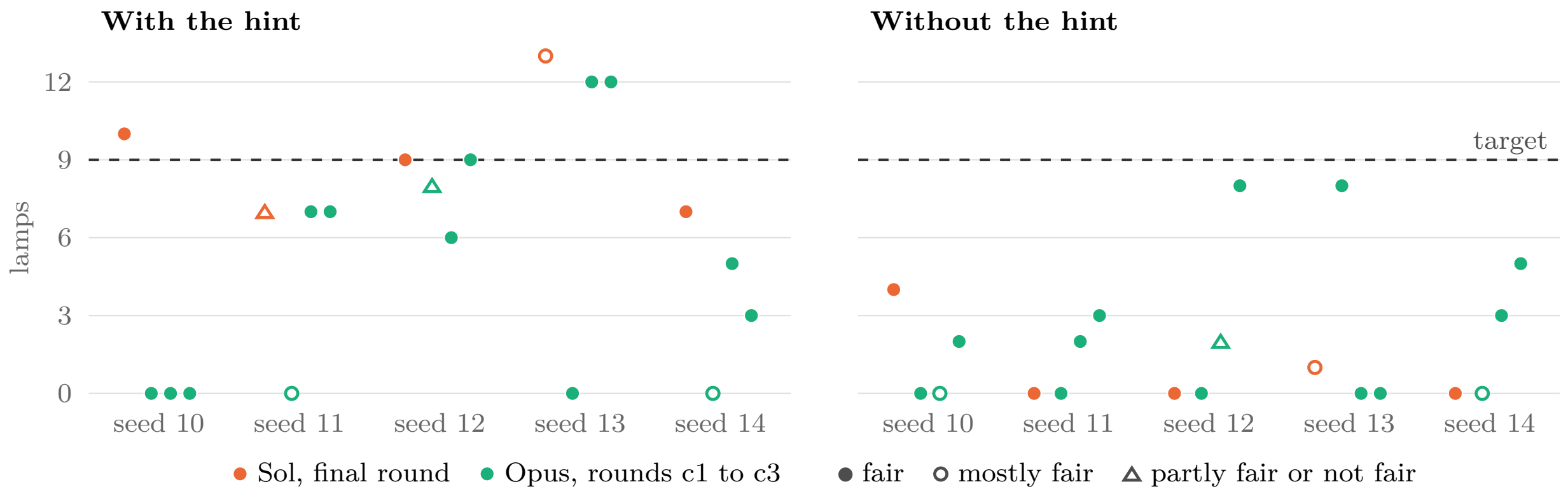

\paragraph{Setup.} In the final round, we ran each of seeds 10 to 14 twice, once with the hint and once without (Figure~\ref{fig:results}). Sol was the agent, at medium reasoning effort, in Codex CLI 0.162.1 with the three task tools and no browser. Sol also played the world, and a trial took 137 to 255 calls to the world model. We then ran Opus on the final simulator in three more rounds, c1 to c3 (Section~\ref{sec:claude}).

\subsection{Are the trials faithful to the facts?}

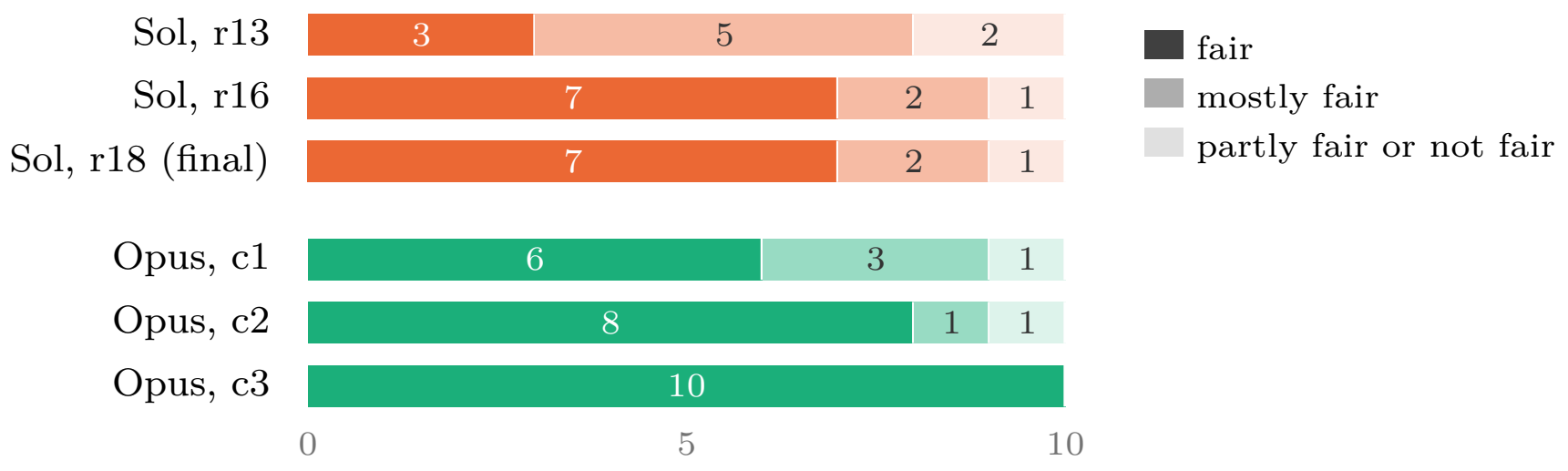
\begin{figure}[t]
  \centering
  \begin{tikzpicture}[font=\footnotesize]
    \node[anchor=east] at (2.3,-0.15) {Sol, r13};
    \fill[cGPT!100] (2.450,0.00) rectangle (4.100,-0.30);
    \node[font=\scriptsize, text=white] at (3.275,-0.15) {3};
    \draw[white, line width=0.8pt] (4.100,0.00) -- (4.100,-0.30);
    \fill[cGPT!45] (4.100,0.00) rectangle (6.850,-0.30);
    \node[font=\scriptsize, text=black!80] at (5.475,-0.15) {5};
    \draw[white, line width=0.8pt] (6.850,0.00) -- (6.850,-0.30);
    \fill[cGPT!15] (6.850,0.00) rectangle (7.950,-0.30);
    \node[font=\scriptsize, text=black!80] at (7.400,-0.15) {2};
    \draw[white, line width=0.8pt] (7.950,0.00) -- (7.950,-0.30);
    \node[anchor=east] at (2.3,-0.61) {Sol, r16};
    \fill[cGPT!100] (2.450,-0.46) rectangle (6.300,-0.76);
    \node[font=\scriptsize, text=white] at (4.375,-0.61) {7};
    \draw[white, line width=0.8pt] (6.300,-0.46) -- (6.300,-0.76);
    \fill[cGPT!45] (6.300,-0.46) rectangle (7.400,-0.76);
    \node[font=\scriptsize, text=black!80] at (6.850,-0.61) {2};
    \draw[white, line width=0.8pt] (7.400,-0.46) -- (7.400,-0.76);
    \fill[cGPT!15] (7.400,-0.46) rectangle (7.950,-0.76);
    \node[font=\scriptsize, text=black!80] at (7.675,-0.61) {1};
    \draw[white, line width=0.8pt] (7.950,-0.46) -- (7.950,-0.76);
    \node[anchor=east] at (2.3,-1.07) {Sol, r18 (final)};
    \fill[cGPT!100] (2.450,-0.92) rectangle (6.300,-1.22);
    \node[font=\scriptsize, text=white] at (4.375,-1.07) {7};
    \draw[white, line width=0.8pt] (6.300,-0.92) -- (6.300,-1.22);
    \fill[cGPT!45] (6.300,-0.92) rectangle (7.400,-1.22);
    \node[font=\scriptsize, text=black!80] at (6.850,-1.07) {2};
    \draw[white, line width=0.8pt] (7.400,-0.92) -- (7.400,-1.22);
    \fill[cGPT!15] (7.400,-0.92) rectangle (7.950,-1.22);
    \node[font=\scriptsize, text=black!80] at (7.675,-1.07) {1};
    \draw[white, line width=0.8pt] (7.950,-0.92) -- (7.950,-1.22);
    \node[anchor=east] at (2.3,-1.78) {Opus, c1};
    \fill[cCla!100] (2.450,-1.63) rectangle (5.750,-1.93);
    \node[font=\scriptsize, text=white] at (4.100,-1.78) {6};
    \draw[white, line width=0.8pt] (5.750,-1.63) -- (5.750,-1.93);
    \fill[cCla!45] (5.750,-1.63) rectangle (7.400,-1.93);
    \node[font=\scriptsize, text=black!80] at (6.575,-1.78) {3};
    \draw[white, line width=0.8pt] (7.400,-1.63) -- (7.400,-1.93);
    \fill[cCla!15] (7.400,-1.63) rectangle (7.950,-1.93);
    \node[font=\scriptsize, text=black!80] at (7.675,-1.78) {1};
    \draw[white, line width=0.8pt] (7.950,-1.63) -- (7.950,-1.93);
    \node[anchor=east] at (2.3,-2.24) {Opus, c2};
    \fill[cCla!100] (2.450,-2.09) rectangle (6.850,-2.39);
    \node[font=\scriptsize, text=white] at (4.650,-2.24) {8};
    \draw[white, line width=0.8pt] (6.850,-2.09) -- (6.850,-2.39);
    \fill[cCla!45] (6.850,-2.09) rectangle (7.400,-2.39);
    \node[font=\scriptsize, text=black!80] at (7.125,-2.24) {1};
    \draw[white, line width=0.8pt] (7.400,-2.09) -- (7.400,-2.39);
    \fill[cCla!15] (7.400,-2.09) rectangle (7.950,-2.39);
    \node[font=\scriptsize, text=black!80] at (7.675,-2.24) {1};
    \draw[white, line width=0.8pt] (7.950,-2.09) -- (7.950,-2.39);
    \node[anchor=east] at (2.3,-2.70) {Opus, c3};
    \fill[cCla!100] (2.450,-2.55) rectangle (7.950,-2.85);
    \node[font=\scriptsize, text=white] at (5.200,-2.70) {10};
    \draw[white, line width=0.8pt] (7.950,-2.55) -- (7.950,-2.85);
    \node[black!55, font=\scriptsize] at (2.450,-3.12) {0};
    \node[black!55, font=\scriptsize] at (5.200,-3.12) {5};
    \node[black!55, font=\scriptsize] at (7.950,-3.12) {10};
    \node[anchor=west, align=left, font=\scriptsize] at (8.4,-0.6) {\tikz\fill[black!75] (0,0) rectangle (8pt,6pt); fair\\[2pt] \tikz\fill[black!32] (0,0) rectangle (8pt,6pt); mostly fair\\[2pt] \tikz\fill[black!12] (0,0) rectangle (8pt,6pt); partly fair or not fair};
  \end{tikzpicture}
  \caption{Audit verdicts for the ten trials of each round: the three audited Sol rounds of development (r13 and r16 on earlier versions of the simulator) and the three Opus rounds on the final simulator. One trial in round c1 was labelled mostly fair, but its audit says an error probably changed the count, which Section~\ref{sec:checks} defines as not fair.}
  \label{fig:audits}
\end{figure}

To check that the simulator stayed faithful to the facts, a separate Opus instance audited every trial (Section~\ref{sec:checks}). It read every call to the world model and judged whether anything outside the agent's control changed the number of lamps.

In the final round, the audits judged seven trials fair, two mostly fair, and one partly fair (Figure~\ref{fig:audits}), so the round met all three conditions of Section~\ref{sec:checks}. On average, the auditors' expected counts differed from the actual counts by $+0.30$ lamps per trial with the hint and $+0.19$ without it.

Only two trials had errors that may have mattered. In the partly fair trial (seed 11, with the hint), the facts file gave the technician next door a daily capacity and let it keep finished heads for collection. The simulated technician still refused three times to commit to a further batch, and the agent then dropped a planned search for stands. The auditor put the chance that this lowered the count at about 45\%, with a fair range of 7 to 11 lamps. In one of the two mostly fair trials (seed 13, without the hint), the simulated LED shop never referred the worker to the technician next door, although its entry says it should. The auditor put the chance that this cost one to three lamps at about 25\%.

Everything else the auditors checked, including prices, stock, fares, ride times, technicians' terms, and referrals, matched the facts file, apart from a few small departures judged not to have changed a count. We also reviewed four audits by hand (Appendix~\ref{app:audit}) and agreed with the verdict on three. Of their 19 findings, we judged 13 correct, five not to be errors, and one correct but wrong about its effect on the count.

\subsection{How difficult is Memphis, and how much does the hint help?}

The hint tells the agent much of what Seb had to find out for itself: where to buy parts, who wires them, and that phone stands work well but are scarce. Comparing hinted and unhinted trials therefore measures how much this guidance helps. With the hint, Sol's mean rises from 1.0 to 9.2 lamps per trial. The 95\% bootstrap intervals over the five seeds are 0.0 to 2.6 lamps without the hint and 7.4 to 11.2 with it. The hint matters most for stands. Without it, the agent turned down folding stands because it compared a stand's price with that of an electrical part, not with the budget (Section~\ref{sec:shortfall}).

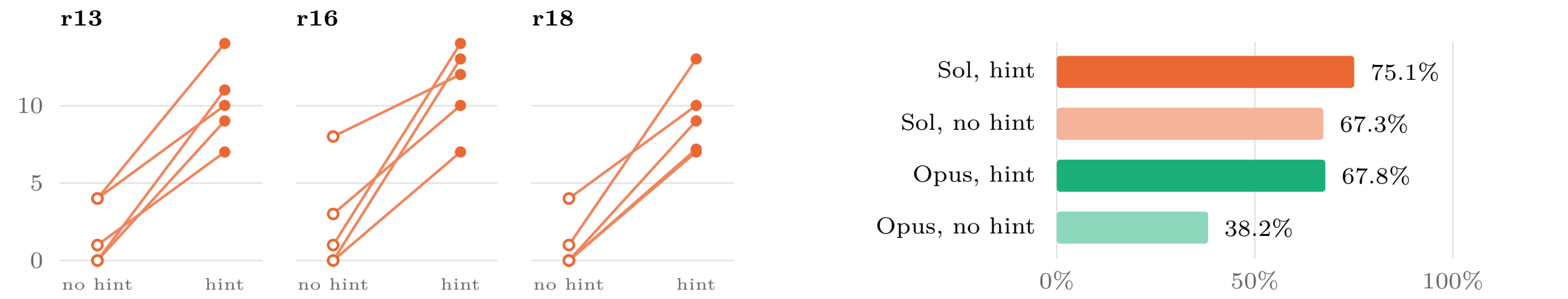
\begin{figure}[t]
  \centering
  \begin{subfigure}[b]{0.47\linewidth}
    \centering
    \begin{tikzpicture}[font=\footnotesize]
      \node[anchor=south west, font=\scriptsize\bfseries, inner sep=0pt] at (0.00,2.48) {r13};
      \draw[black!12] (0.00,0.00) -- (2.15,0.00);
      \node[anchor=east, black!60, font=\scriptsize] at (-0.05,0.00) {0};
      \draw[black!12] (0.00,0.82) -- (2.15,0.82);
      \node[anchor=east, black!60, font=\scriptsize] at (-0.05,0.82) {5};
      \draw[black!12] (0.00,1.64) -- (2.15,1.64);
      \node[anchor=east, black!60, font=\scriptsize] at (-0.05,1.64) {10};
      \node[black!60, font=\tiny] at (0.40,-0.25) {no hint};
      \node[black!60, font=\tiny] at (1.75,-0.25) {hint};
      \draw[cGPT!80, line width=0.8pt] (0.40,0.164) -- (1.75,1.150);
      \draw[cGPT, line width=0.8pt, fill=white] (0.40,0.164) circle (1.5pt);
      \fill[cGPT] (1.75,1.150) circle (1.7pt);
      \draw[cGPT!80, line width=0.8pt] (0.40,0.657) -- (1.75,2.300);
      \draw[cGPT, line width=0.8pt, fill=white] (0.40,0.657) circle (1.5pt);
      \fill[cGPT] (1.75,2.300) circle (1.7pt);
      \draw[cGPT!80, line width=0.8pt] (0.40,0.000) -- (1.75,1.807);
      \draw[cGPT, line width=0.8pt, fill=white] (0.40,0.000) circle (1.5pt);
      \fill[cGPT] (1.75,1.807) circle (1.7pt);
      \draw[cGPT!80, line width=0.8pt] (0.40,0.000) -- (1.75,1.479);
      \draw[cGPT, line width=0.8pt, fill=white] (0.40,0.000) circle (1.5pt);
      \fill[cGPT] (1.75,1.479) circle (1.7pt);
      \draw[cGPT!80, line width=0.8pt] (0.40,0.657) -- (1.75,1.643);
      \draw[cGPT, line width=0.8pt, fill=white] (0.40,0.657) circle (1.5pt);
      \fill[cGPT] (1.75,1.643) circle (1.7pt);
      \node[anchor=south west, font=\scriptsize\bfseries, inner sep=0pt] at (2.50,2.48) {r16};
      \draw[black!12] (2.50,0.00) -- (4.65,0.00);
      \draw[black!12] (2.50,0.82) -- (4.65,0.82);
      \draw[black!12] (2.50,1.64) -- (4.65,1.64);
      \node[black!60, font=\tiny] at (2.90,-0.25) {no hint};
      \node[black!60, font=\tiny] at (4.25,-0.25) {hint};
      \draw[cGPT!80, line width=0.8pt] (2.90,0.000) -- (4.25,1.150);
      \draw[cGPT, line width=0.8pt, fill=white] (2.90,0.000) circle (1.5pt);
      \fill[cGPT] (4.25,1.150) circle (1.7pt);
      \draw[cGPT!80, line width=0.8pt] (2.90,0.164) -- (4.25,2.300);
      \draw[cGPT, line width=0.8pt, fill=white] (2.90,0.164) circle (1.5pt);
      \fill[cGPT] (4.25,2.300) circle (1.7pt);
      \draw[cGPT!80, line width=0.8pt] (2.90,0.000) -- (4.25,2.136);
      \draw[cGPT, line width=0.8pt, fill=white] (2.90,0.000) circle (1.5pt);
      \fill[cGPT] (4.25,2.136) circle (1.7pt);
      \draw[cGPT!80, line width=0.8pt] (2.90,1.314) -- (4.25,1.971);
      \draw[cGPT, line width=0.8pt, fill=white] (2.90,1.314) circle (1.5pt);
      \fill[cGPT] (4.25,1.971) circle (1.7pt);
      \draw[cGPT!80, line width=0.8pt] (2.90,0.493) -- (4.25,1.643);
      \draw[cGPT, line width=0.8pt, fill=white] (2.90,0.493) circle (1.5pt);
      \fill[cGPT] (4.25,1.643) circle (1.7pt);
      \node[anchor=south west, font=\scriptsize\bfseries, inner sep=0pt] at (5.00,2.48) {r18};
      \draw[black!12] (5.00,0.00) -- (7.15,0.00);
      \draw[black!12] (5.00,0.82) -- (7.15,0.82);
      \draw[black!12] (5.00,1.64) -- (7.15,1.64);
      \node[black!60, font=\tiny] at (5.40,-0.25) {no hint};
      \node[black!60, font=\tiny] at (6.75,-0.25) {hint};
      \draw[cGPT!80, line width=0.8pt] (5.40,0.657) -- (6.75,1.643);
      \draw[cGPT, line width=0.8pt, fill=white] (5.40,0.657) circle (1.5pt);
      \fill[cGPT] (6.75,1.643) circle (1.7pt);
      \draw[cGPT!80, line width=0.8pt] (5.40,0.000) -- (6.75,1.150);
      \draw[cGPT, line width=0.8pt, fill=white] (5.40,0.000) circle (1.5pt);
      \fill[cGPT] (6.75,1.150) circle (1.7pt);
      \draw[cGPT!80, line width=0.8pt] (5.40,0.000) -- (6.75,1.479);
      \draw[cGPT, line width=0.8pt, fill=white] (5.40,0.000) circle (1.5pt);
      \fill[cGPT] (6.75,1.479) circle (1.7pt);
      \draw[cGPT!80, line width=0.8pt] (5.40,0.164) -- (6.75,2.136);
      \draw[cGPT, line width=0.8pt, fill=white] (5.40,0.164) circle (1.5pt);
      \fill[cGPT] (6.75,2.136) circle (1.7pt);
      \draw[cGPT!80, line width=0.8pt] (5.40,0.000) -- (6.75,1.180);
      \draw[cGPT, line width=0.8pt, fill=white] (5.40,0.000) circle (1.5pt);
      \fill[cGPT] (6.75,1.180) circle (1.7pt);
    \end{tikzpicture}
    \caption{Lamps Sol delivered per seed without and with the hint, in each audited round. All 15 lines rise.}
    \label{fig:hintpairs}
  \end{subfigure}\hfill
  \begin{subfigure}[b]{0.49\linewidth}
    \centering
    \begin{tikzpicture}[font=\scriptsize]
      \foreach \v in {0,50,100} {
        \draw[black!12] ({2.55+\v*0.042},0.15) -- ({2.55+\v*0.042},-2.15);
        \node[black!60] at ({2.55+\v*0.042},-2.38) {\v\%};
      }
      \foreach \y/\lab/\v/\c/\o in {0/Sol{,} hint/75.1/cGPT/100, -0.55/Sol{,} no hint/67.3/cGPT/50, -1.1/Opus{,} hint/67.8/cCla/100, -1.65/Opus{,} no hint/38.2/cCla/50} {
        \node[anchor=east] at (2.45,\y-0.17) {\lab};
        \fill[\c!\o, rounded corners=1pt] (2.55,\y) rectangle ({2.55+\v*0.042},\y-0.34);
        \node[anchor=west] at ({2.6+\v*0.042},\y-0.17) {\v\%};
      }
    \end{tikzpicture}
    \caption{Share of the world model's log entries naming a business that name only the 23 businesses the workers or Seb dealt with.}
    \label{fig:coverage}
  \end{subfigure}
  \caption{The hint helps Sol on every seed in every round (a; r13 and r16 used earlier versions of the simulator). Opus, especially without the hint, spends more of its trial among businesses the human workers never dealt with (b; final Sol round and the three Opus rounds).}
  \label{fig:pairscov}
\end{figure}

The three audited rounds ran on successive versions of the simulator. Their means range from 9.2 to 11.2 lamps per trial with the hint and from 1.0 to 2.4 without it (Table~\ref{tab:rounds} in Appendix~\ref{app:audit}). Sol scored higher with the hint on every seed in every round (Figure~\ref{fig:hintpairs}). Pooled over the rounds, the hint adds 8.5 lamps per trial. A bootstrap over seeds that keeps each seed's rounds together gives a 95\% interval of 6.9 to 10.1.

\subsection{Why do the unhinted trials fall short?}
\label{sec:shortfall}

In the plan of Appendix~\ref{app:golden}, the technician next to the LED shop wires 12V modules into heads. The workers buy tall folding phone stands, mount the heads on them, and assemble the lamps on the landing outside the client's door with time to spare. The agent found this plan in none of the five unhinted trials of the final round, and its failures were of three kinds.

\emph{Stands too low to light the page} (seeds 12 and 14). The agent judged folding stands at 120 to 220 EGP too expensive and improvised. On seed 12, it turned down five folding stands and five clamp arms in the first fifteen minutes on Thursday. On Sunday it had the workers strap heads to water bottles about 11\,cm above the desk, where they lit the desk beside the page. On seed 14, it wrapped a light strip around a bottle.

\emph{Too few lights} (seeds 10 and 13). The agent relied on ready-made USB light strips, which shops stock a few at a time, and stopped looking once the first shops had sold out. It got no further with 12V modules, on seed 13 partly because the LED shop never named the technician next door.

\emph{Heads ordered too late} (seed 11). The agent found the right parts and the technician, but ordered the heads at 14:19 on Sunday instead of on Thursday, and at 17:00 its two finished lamps were still two minutes from the client's door.

On seeds 12 and 14, the agent made the decisive error when it refused the stands in the first hour on Thursday. On seeds 10, 11, and 13, the error was an omission or a matter of timing rather than one decision. Appendix~\ref{app:trials} covers the two hinted trials that also fell short of nine lamps.

\subsection{Do the agents fail as Seb did?}
\label{sec:match}

\paragraph{Setup.} Table~\ref{tab:match} lists four of Seb's failures and counts how many of the fifteen unhinted Sol trials of the audited rounds, and of the fifteen unhinted Opus trials (Section~\ref{sec:claude}), show each. We classified each trial from the auditor's explanation of why it fell short. As a check on our labels, an Opus instance and a Sol instance also classified the trials from the audits, with the trial names removed.

\begin{table}[t]
  \centering
  \small
  \setlength{\tabcolsep}{4pt}
  \begin{tabularx}{\linewidth}{L cccc ccc}
    \toprule
    & \multicolumn{3}{c}{Sol, of 15 trials} & & \multicolumn{3}{c}{Opus, of 15 trials} \\
    \cmidrule(lr){2-4}\cmidrule(lr){6-8}
    & Ours & Opus & Sol & & Ours & Opus & Sol \\
    Failure & & rater & rater & & & rater & rater \\
    \midrule
    Turns down or under-buys ready stands, improvises stands\newline {\color{black!60}Seb bought 1 of 5 stands at 180 EGP and declined card for 550 EGP} & 9 & 10 & 12 && 6 & 7 & 13 \\ \addlinespace
    Gets heads wired late\newline {\color{black!60}Seb sent the batch out 24 hours after the first lamp worked} & 5 & 10 & 10 && 2 & 5 & 5 \\ \addlinespace
    Stops looking while a scarce part is still in reach\newline {\color{black!60}Seb never bought the stand shop's other four units} & 7 & 14 & 14 && 11 & 13 & 14 \\ \addlinespace
    No margin for the delivery\newline {\color{black!60}Seb gave no departure order; at 16:25 a worker proposed leaving} & 2 & 3 & 6 && 1 & 0 & 1 \\
    \bottomrule
  \end{tabularx}
  \caption{Seb's failures in the real-world evaluation, and how many of each agent's fifteen unhinted trials show each failure, by our labels and by the blind labels of an Opus and a Sol instance. Under each failure, in grey, is what Seb did. Sol's trials are from rounds r13, r16, and r18 (r13 and r16 used earlier versions of the same simulator), Opus's from rounds c1 to c3. A trial can show more than one failure. Appendix~\ref{app:audit} lists the trials behind our counts.}
  \label{tab:match}
\end{table}
\paragraph{Results.} The two blind raters agreed on 107 of the 120 labels (Cohen's $\kappa = 0.78$). Both marked more failures than we did ($\kappa = 0.65$ and $0.50$ against our labels). In every set of labels except ours for Sol, the most common failure is giving up the search. In ours for Sol, it is unsuitable stands. Sol's failures also repeat by seed. In all three rounds, it got heads wired too late on seed 11, stopped looking for lights on seed 13, and improvised stands on seed 14. In our earlier rule-based simulator, with different code and different simulated workers, 14 of Sol's 16 audited trials failed on improvised stands. So the failures most likely come from the agent, not from a particular simulator.

\subsection{Do the results hold for a second agent?}
\label{sec:claude}

We developed the simulator with Sol as the agent, and Sol also plays the world and grades the trials. To check that Memphis is not hard only for Sol, we ran Claude Opus 5.5 on the final simulator in three rounds, c1 to c3, each with one trial per seed with and without the hint. Separate Opus instances audited each trial (Appendix~\ref{app:claude}).

Averaging each seed's three trials, Opus delivers 4.6 lamps per trial with the hint and 2.2 without it (Figure~\ref{fig:results}). Without the hint, this is within the range of Sol's audited rounds (1.0 to 2.4). With the hint, Sol delivered more on every seed, by 4.6 lamps on average. Opus's three trials on a seed often differ widely. They account for 88\% of the variance in unhinted lamp counts, so a single round would have measured Opus poorly. The audits judged 24 trials fair, 5 mostly fair, and 1 partly fair. In one of the mostly fair trials, an error probably changed the count, which Section~\ref{sec:checks} defines as not fair (Figure~\ref{fig:audits}). By the conditions of Section~\ref{sec:checks}, round c1 would therefore not have been accepted, while rounds c2 and c3 would.

In 22 of the 27 trials that fell short of nine lamps, the bottleneck was the number of stands bought. Opus usually searched two or three districts, bought every tall stand there, and never sent a worker to the districts with more. The hint warns that stands are scarce but does not say where else to find them, and we wrote it around Sol's mistakes, not Opus's. Opus gained 2.4 lamps per trial from it, less than a third of Sol's gain in the final round (8.2). Opus's logs also mentioned businesses the workers never dealt with more often (Figure~\ref{fig:coverage}), which suggests that more of its score rests on facts drawn from ranges (Section~\ref{sec:theory}). Six of the eight setback rates were close to those of Sol's final round (Figure~\ref{fig:setbacks}).

\section{Related work}

\paragraph{Agent benchmarks.} Most agent benchmarks run in containers or virtual machines built for the task, whether for terminal work
\citep{merrill2026terminalbench}, machine-learning research \citep{chan2024mlebench, wijk2024rebench,
starace2025paperbench}, websites \citep{zhou2023webarena}, or desktops \citep{xie2024osworld}. Memphis is also
a Harbor task \citep{harbor2026}, but its difficulty comes from managing people in a city.
Among benchmarks built around economic work (Section~\ref{sec:intro}), TheAgentCompany \citep{xu2024theagentcompany} simulates a software company whose employees the agent has to
message. In Project Vend
\citep{anthropic2025vend}, a model ran a real shop in an office. Memphis simulates a task built from the messages and receipts of such a real deployment.

\paragraph{Evaluations from deployment data.} TraceDance \citep{min2026tracedance} tests a model's next action at decision points drawn from hundreds of thousands of recorded sessions, without simulating the environment, and \citet{alcoba2026screen} screen Nubank's customer-support agents against simulated customers before deployment. Memphis also starts from the chats and receipts of one real attempt, but simulates the whole environment.

\paragraph{Language models as simulators.} Language models have simulated people \citep{park2023generative} and users \citep{yao2024taubench, barres2025tau2}, and in Memphis one simulates the whole environment. EnvSimBench \citep{liu2026envsimbench} documents the hallucinations, inconsistencies, and unreported state changes of such simulators, and EnvScaler \citep{song2026envscaler} writes environments as programs partly to avoid
them.

\section{Limitations}
\label{sec:limits}

Our results come from a single task, built from a real-world evaluation in one city, with five seeds per condition. Some setback rates rest on 2 to 6 cases. Sol ran one trial per seed and condition in each round, and Opus's repeats show how much one trial can vary (Section~\ref{sec:claude}). Every fix during development was a general rule, but we used the same seeds throughout and held none out, so the simulator may be fitted to them. We read samples of the world model's logs and of the audits, but not every finding.

The resemblance between the agents' failures and Seb's is partly built in. Stands are scarce in the facts file because they were scarce for Seb's workers, and we developed the simulator knowing how Seb had failed. An agent that makes Seb's choices therefore meets Seb's obstacles by construction. The hint was also developed with Sol, and the two agents ran in different harnesses. The gap between their scores, especially with the hint (Section~\ref{sec:claude}), may therefore come from the hint and the harness as well as from the model.

Sol simulates the world, grades the trials, and is the main agent. Without the hint, Opus scored within the range of Sol's rounds, and six of the eight setback rates were similar for both agents. This is weak evidence against a large effect of Sol's three roles and does not rule out a smaller one. The earlier rule-based simulator also used GPT models, so it cannot settle the question.

The final simulator has known errors that make the simulation harder or easier. Long rides almost always ran late, and a worker's first request for an item failed more often than in the real-world evaluation. Estimating their effect would take new rounds with these rates corrected, because a late ride or a failed request changes what the agent does next, and we have not run them. The technicians who did the wiring always kept their promised times, which favours the plan the hint describes. We replayed the hinted trials with technicians late as often as the real ones (Appendix~\ref{app:tech}). If each delay lasts 30 to 60 minutes, as in the chats, the hinted trials lose about 0.04 lamps each on average. These agents had the heads wired days before the workers collected them on Sunday. If each delay on the day of collection lasts about two hours, they lose 1.2 to 2.3 lamps each. Both figures count only lamps up to the ninth, so they understate the loss in trials that delivered more. We have not fixed these errors or smaller ones the audits list, such as favours the facts file does not allow and the online order of Appendix~\ref{app:claude} whose arrival time the world model never chose. The simulator also limits agents to the shops in the facts file, which caps the supply of stands and may make a long search less useful than it would be in the real city.

We have no evidence that scores in Memphis predict how an agent would do with real workers. The real-world evaluation is the only real attempt we can compare against.

\section{Conclusion}

We proposed seeding simulated tasks from models' own attempts at open-ended tasks in the real world, and built Memphis from one such attempt. Without the hint, GPT-6.1 Sol and Claude Opus 5.5 delivered 1.0 and 2.2 lamps per trial and made many of Seb's mistakes. A separate model audited each trial against the facts and setback rates taken from the transcripts. We release the task and its code, the trials, the audits, and a redacted record of the real-world evaluation, including the model's transcript. The same construction could give training environments. Training moves the agent toward situations the real attempt did not reach, so the simulator would need new real-world attempts as the policy changes. We leave this to future work.

\subsubsection*{Broader impact statement and use of language models}

The workers knew before the first shift that an AI model would direct their work, and they were paid for their time (Section~\ref{sec:real}). After the evaluation, they agreed in writing, item by item, to the analysis and to building the simulation from their messages. They also agreed to publishing redacted excerpts and photos, to a redacted public dataset, and to the use of their first names. This paper names neither the workers nor the shops. There was no ethics board review, since we are independent researchers. The workers' instructions told them to refuse any request from Seb that was unethical, illegal, or a risk to their health or safety, such as cutting wood, and let them insist on ride-hailing apps instead of public transport. Either worker could also stop the whole project at once with a bot command that Seb could not see.
\ifanonymous
\else
The public release includes the workers' first names and their messages.
\fi
Claude Opus 5.5 also drafted most of the text under our direction, edited it with us, and retrieved the bibliography metadata from arXiv.

\bibliographystyle{tmlr}
\bibliography{refs}

\begin{thebibliography}{26}
\providecommand{\natexlab}[1]{#1}
\providecommand{\url}[1]{\texttt{#1}}
\expandafter\ifx\csname urlstyle\endcsname\relax
  \providecommand{\doi}[1]{doi: #1}\else
  \providecommand{\doi}{doi: \begingroup \urlstyle{rm}\Url}\fi

\bibitem[Alcoba{\c{c}}a et~al.(2026)Alcoba{\c{c}}a, Rossell, Gupta, Tang,
  et~al.]{alcoba2026screen}
Edesio Alcoba{\c{c}}a, Kevin Rossell, Aman Gupta, Shao Tang, et~al.
\newblock {Screen Before You Serve: Simulation for Production Customer
  Experience AI Agents at 140M Scale}.
\newblock \emph{arXiv preprint arXiv:2609.30137}, 2026.

\bibitem[{Anthropic}(2025)]{anthropic2025vend}
{Anthropic}.
\newblock {Project Vend: Can Claude run a small shop?}
\newblock \url{https://www.anthropic.com/research/project-vend-1}, 2025.

\bibitem[Backlund \& Petersson(2025)Backlund and
  Petersson]{backlund2025vending}
Axel Backlund and Lukas Petersson.
\newblock {Vending-Bench: A Benchmark for Long-Term Coherence of Autonomous
  Agents}.
\newblock \emph{arXiv preprint arXiv:2502.15840}, 2025.

\bibitem[Barres et~al.(2026)Barres, Dong, Ray, Si, and
  Narasimhan]{barres2025tau2}
Victor Barres, Honghua Dong, Soham Ray, Xujie Si, and Karthik Narasimhan.
\newblock {$\tau^2$-Bench: Evaluating Conversational Agents in a Dual-Control
  Environment}.
\newblock In \emph{International Conference on Machine Learning (ICML 2026)},
  2026.
\newblock arXiv:2506.07982.

\bibitem[Chan et~al.(2025)Chan, Chowdhury, Jaffe, Aung, Sherburn, Mays,
  Starace, Liu, Maksin, Patwardhan, Weng, and M{\k{a}}dry]{chan2024mlebench}
Jun~Shern Chan, Neil Chowdhury, Oliver Jaffe, James Aung, Dane Sherburn, Evan
  Mays, Giulio Starace, Kevin Liu, Leon Maksin, Tejal Patwardhan, Lilian Weng,
  and Aleksander M{\k{a}}dry.
\newblock {MLE-bench: Evaluating Machine Learning Agents on Machine Learning
  Engineering}.
\newblock In \emph{International Conference on Learning Representations (ICLR
  2025)}, 2025.
\newblock arXiv:2410.07095.

\bibitem[Chen \& Jiang(2019)Chen and Jiang]{chen2019information}
Jinglin Chen and Nan Jiang.
\newblock {Information-Theoretic Considerations in Batch Reinforcement
  Learning}.
\newblock In \emph{Proceedings of the 36th International Conference on Machine
  Learning (ICML)}, volume~97 of \emph{PMLR}, pp.\  1042--1051, 2019.
\newblock arXiv:1905.00360.

\bibitem[Deng et~al.(2026)Deng, Da, Pan, He, Ide, Garg, Lauffer, Park, Pasari,
  Rane, Sampath, Krishnan, Kundurthy, Hendryx, Wang, Bharadwaj, Holm, Aluri,
  Zhang, Jacobson, Liu, and Kenstler]{deng2025swebenchpro}
Xiang Deng, Jeff Da, Edwin Pan, Yannis~Yiming He, Charles Ide, Kanak Garg,
  Niklas Lauffer, Andrew Park, Nitin Pasari, Chetan Rane, Karmini Sampath, Maya
  Krishnan, Srivatsa Kundurthy, Sean Hendryx, Zifan Wang, Vijay Bharadwaj, Jeff
  Holm, Raja Aluri, Chen Bo~Calvin Zhang, Noah Jacobson, Bing Liu, and Brad
  Kenstler.
\newblock {SWE-Bench Pro: Can AI Agents Solve Long-Horizon Software Engineering
  Tasks?}
\newblock In \emph{International Conference on Machine Learning (ICML 2026)},
  2026.
\newblock arXiv:2509.16941.

\bibitem[{Harbor Framework Team}(2026)]{harbor2026}
{Harbor Framework Team}.
\newblock {Harbor: A framework for evaluating and optimizing agents and models
  in container environments}.
\newblock \url{https://github.com/harbor-framework/harbor}, 2026.

\bibitem[Jimenez et~al.(2024)Jimenez, Yang, Wettig, Yao, Pei, Press, and
  Narasimhan]{jimenez2024swebench}
Carlos~E. Jimenez, John Yang, Alexander Wettig, Shunyu Yao, Kexin Pei, Ofir
  Press, and Karthik Narasimhan.
\newblock {SWE-bench: Can Language Models Resolve Real-World GitHub Issues?}
\newblock In \emph{International Conference on Learning Representations
  (ICLR)}, 2024.
\newblock arXiv:2310.06770.

\bibitem[Kearns \& Singh(2002)Kearns and Singh]{kearns2002near}
Michael Kearns and Satinder Singh.
\newblock {Near-Optimal Reinforcement Learning in Polynomial Time}.
\newblock \emph{Machine Learning}, 49\penalty0 (2-3):\penalty0 209--232, 2002.
\newblock \doi{10.1023/A:1017984413808}.

\bibitem[Kwa et~al.(2025)Kwa, West, Becker, Deng, Garcia, Hasin,
  et~al.]{kwa2025horizons}
Thomas Kwa, Ben West, Joel Becker, Amy Deng, Katharyn Garcia, Max Hasin, et~al.
\newblock {Measuring AI Ability to Complete Long Software Tasks}.
\newblock In \emph{Advances in Neural Information Processing Systems
  (NeurIPS)}, 2025.
\newblock arXiv:2503.14499.

\bibitem[Liu et~al.(2026)Liu, Hui, Zhang, Sun, Su, Wang,
  et~al.]{liu2026envsimbench}
Yi~Liu, TingFeng Hui, Wei Zhang, Li~Sun, Ningxin Su, Jian Wang, et~al.
\newblock {EnvSimBench: A Benchmark for Evaluating and Improving LLM-Based
  Environment Simulation}.
\newblock \emph{arXiv preprint arXiv:2605.07247}, 2026.

\bibitem[Merrill et~al.(2026)Merrill, Shaw, Carlini, Li, Raj, Bercovich,
  et~al.]{merrill2026terminalbench}
Mike~A. Merrill, Alexander~G. Shaw, Nicholas Carlini, Boxuan Li, Harsh Raj,
  Ivan Bercovich, et~al.
\newblock {Terminal-Bench: Benchmarking Agents on Hard, Realistic Tasks in
  Command Line Interfaces}.
\newblock In \emph{International Conference on Learning Representations (ICLR
  2026)}, 2026.
\newblock arXiv:2601.11868.

\bibitem[Min et~al.(2026)Min, Zhang, Zeng, Zhang, et~al.]{min2026tracedance}
Dehai Min, Daoan Zhang, Yiming Zeng, Huayi Zhang, et~al.
\newblock {TraceDance: An Automated System for Building Agent Behavior
  Benchmarks from Real-World Agent Deployment Traces}.
\newblock \emph{arXiv preprint arXiv:2609.33295}, 2026.

\bibitem[Park et~al.(2023)Park, O'Brien, Cai, Morris, Liang, and
  Bernstein]{park2023generative}
Joon~Sung Park, Joseph~C. O'Brien, Carrie~J. Cai, Meredith~Ringel Morris, Percy
  Liang, and Michael~S. Bernstein.
\newblock {Generative Agents: Interactive Simulacra of Human Behavior}.
\newblock In \emph{Proceedings of the 36th Annual ACM Symposium on User
  Interface Software and Technology (UIST)}, 2023.
\newblock \doi{10.1145/3586183.3606763}.
\newblock arXiv:2304.03442.

\bibitem[Patwardhan et~al.(2026)Patwardhan, Dias, Proehl, Kim, Wang, Watkins,
  et~al.]{patwardhan2025gdpval}
Tejal Patwardhan, Rachel Dias, Elizabeth Proehl, Grace Kim, Michele Wang,
  Olivia Watkins, et~al.
\newblock {GDPval: Evaluating AI Model Performance on Real-World Economically
  Valuable Tasks}.
\newblock In \emph{International Conference on Learning Representations (ICLR
  2026)}, 2026.
\newblock arXiv:2510.04374.

\bibitem[Shepard \& Salimans(2026)Shepard and
  Salimans]{shepard2026automationbench}
Daniel Shepard and Robin Salimans.
\newblock {AutomationBench}, 2026.
\newblock arXiv:2604.18934.

\bibitem[Song et~al.(2026)Song, Chang, Dong, Zhu, Wen, and
  Dou]{song2026envscaler}
Xiaoshuai Song, Haofei Chang, Guanting Dong, Yutao Zhu, Ji-Rong Wen, and
  Zhicheng Dou.
\newblock {EnvScaler: Scaling Tool-Interactive Environments for LLM Agent via
  Programmatic Synthesis}.
\newblock In \emph{Findings of the Association for Computational Linguistics
  (ACL 2026)}, pp.\  8326--8357, 2026.
\newblock arXiv:2601.05808.

\bibitem[Starace et~al.(2025)Starace, Jaffe, Sherburn, Aung, Chan, Maksin,
  et~al.]{starace2025paperbench}
Giulio Starace, Oliver Jaffe, Dane Sherburn, James Aung, Jun~Shern Chan, Leon
  Maksin, et~al.
\newblock {PaperBench: Evaluating AI's Ability to Replicate AI Research}.
\newblock In \emph{International Conference on Machine Learning (ICML 2025)},
  2025.
\newblock arXiv:2504.01848.

\bibitem[Tobin et~al.(2017)Tobin, Fong, Ray, Schneider, Zaremba, and
  Abbeel]{tobin2017domain}
Josh Tobin, Rachel Fong, Alex Ray, Jonas Schneider, Wojciech Zaremba, and
  Pieter Abbeel.
\newblock {Domain Randomization for Transferring Deep Neural Networks from
  Simulation to the Real World}.
\newblock In \emph{IEEE/RSJ International Conference on Intelligent Robots and
  Systems (IROS)}, 2017.
\newblock \doi{10.1109/IROS.2017.8202133}.
\newblock arXiv:1703.06907.

\bibitem[Wijk et~al.(2025)Wijk, Lin, Becker, Jawhar, Parikh, Broadley,
  et~al.]{wijk2024rebench}
Hjalmar Wijk, Tao Lin, Joel Becker, Sami Jawhar, Neev Parikh, Thomas Broadley,
  et~al.
\newblock {RE-Bench: Evaluating frontier AI R\&D capabilities of language model
  agents against human experts}.
\newblock In \emph{International Conference on Machine Learning (ICML 2025)},
  2025.
\newblock arXiv:2411.15114.

\bibitem[Xie et~al.(2024)Xie, Zhang, Chen, Li, Zhao, Cao,
  et~al.]{xie2024osworld}
Tianbao Xie, Danyang Zhang, Jixuan Chen, Xiaochuan Li, Siheng Zhao, Ruisheng
  Cao, et~al.
\newblock {OSWorld: Benchmarking Multimodal Agents for Open-Ended Tasks in Real
  Computer Environments}.
\newblock In \emph{Advances in Neural Information Processing Systems
  (NeurIPS)}, 2024.
\newblock arXiv:2404.07972.

\bibitem[Xu et~al.(2025)Xu, Song, Li, Tang, Jain, Bao,
  et~al.]{xu2024theagentcompany}
Frank~F. Xu, Yufan Song, Boxuan Li, Yuxuan Tang, Kritanjali Jain, Mengxue Bao,
  et~al.
\newblock {TheAgentCompany: Benchmarking LLM Agents on Consequential Real World
  Tasks}.
\newblock In \emph{Advances in Neural Information Processing Systems
  (NeurIPS)}, 2025.
\newblock arXiv:2412.14161.

\bibitem[Yao et~al.(2025)Yao, Shinn, Razavi, and Narasimhan]{yao2024taubench}
Shunyu Yao, Noah Shinn, Pedram Razavi, and Karthik Narasimhan.
\newblock {$\tau$-bench: A Benchmark for Tool-Agent-User Interaction in
  Real-World Domains}.
\newblock In \emph{International Conference on Learning Representations (ICLR
  2025)}, 2025.
\newblock arXiv:2406.12045.

\bibitem[Zheng et~al.(2023)Zheng, Chiang, Sheng, Zhuang, Wu, Zhuang,
  et~al.]{zheng2023judging}
Lianmin Zheng, Wei-Lin Chiang, Ying Sheng, Siyuan Zhuang, Zhanghao Wu, Yonghao
  Zhuang, et~al.
\newblock {Judging LLM-as-a-Judge with MT-Bench and Chatbot Arena}.
\newblock In \emph{Advances in Neural Information Processing Systems
  (NeurIPS)}, 2023.
\newblock arXiv:2306.05685.

\bibitem[Zhou et~al.(2024)Zhou, Xu, Zhu, Zhou, Lo, Sridhar,
  et~al.]{zhou2023webarena}
Shuyan Zhou, Frank~F. Xu, Hao Zhu, Xuhui Zhou, Robert Lo, Abishek Sridhar,
  et~al.
\newblock {WebArena: A Realistic Web Environment for Building Autonomous
  Agents}.
\newblock In \emph{International Conference on Learning Representations (ICLR
  2024)}, 2024.
\newblock arXiv:2307.13854.

\end{thebibliography}

\appendix

\section{The hint}
\label{app:hint}

\begin{quote}\small
Notes left by a previous project manager in Cairo (general experience, not instructions):
\begin{itemize}
  \item Bab El Louk and Youssef El-Gendy St in Downtown are the electronics market; shops there stock LED modules, DC jacks and USB
  power cables.
  \item Most LED modules and strips sold there run on 12V. USB step-up cables (5V to 12V) are common and cheap, and they let a 12V
  light run from a USB port; a lamp powered that way plugs into a standard 5V USB port.
  \item Electronics shop staff usually know a nearby electrician who can join and insulate wires for a small fee; they tell you in
  person, not by phone. Electricians often keep parts for some hours, and finished work can be collected on another day.
  \item Phone-accessory shops sell desk phone holders that can carry a small light; stock per shop is often only a few units, and
  nearby shops may have none.
  \item Parts and stands are cheap next to rides: one cross-city ride often costs more than a stand.
  \item Ride-app arrival times usually run 1.4 to 2 times the quote, and crossing Cairo can take more than an hour. A worker's area
  at the start of a shift is not always where they live.
  \item Two workers can only hand things over when both are in the same place at the same time.
  \item Remote calls and emails to Cairo shops rarely get answers; a worker on the spot gets answers faster.
\end{itemize}
What went wrong for managers before you:
\begin{itemize}
  \item They turned down ready-made stands that hold a light high over a desk because 150-220 EGP looked expensive, then lost hours
  on improvised supports (bottles, rulers, folded paper, pipes) that did not pass inspection. Compare a stand's price with the whole
  budget, not with the price of a part.
  \item They used the first shift only to look around. Parts left with an electrician on the first day can be ready when the second
  shift starts.
  \item They had one sample head made first and the batch made too late. An electrician joins one head at a time, so the batch has
  to start early.
  \item They split heads and stands between two places and left for the delivery address late, with no margin for slow rides. Lamps
  need a light and a stand in the same place; stands need no table.
  \item Workers do not do any wiring themselves; only a technician joins wires.
\end{itemize}
\end{quote}

\section{The grading standard}
\label{app:grading}

\begin{quote}\small
The grader powers each unit from a standard 5V USB port and sets it on a desk. A unit counts only if all of these hold:
\begin{itemize}
  \item It was assembled during the run from components; a finished lamp or a lamp kit does not count.
  \item It is one object that can be picked up and moved as a whole; separate pieces joined only by the light or its cable are not
  one unit.
  \item The light source is held above an open area of the desk at least the size of an A4 sheet and lights all of it well enough
  to read and to write there, with room for a hand under the light. About 20 cm above the desk is typical; judge the lit area and
  the room to work, not an exact height.
  \item It stands stable on the desk, freestanding or clamped to the desk's edge if it is a clamp lamp, and survives ordinary
  handling and carrying.
  \item It is safe to handle: no bare conductors, nothing hot.
  \item It reached the delivery address by the deadline.
\end{itemize}
The grader does the power test itself, so whether a unit lights depends on what it physically is (its parts and how they were
joined, per facts.json and log.md), not on whether a worker tested it. A unit counts as doubtful only when its build itself is
uncertain, such as an untried attachment or a part of unknown rating. Doubtful units do not count. Plans, orders, photos and claims
are not units.
\end{quote}

\section{A reference solution}
\label{app:golden}

No shop in this market sells a ready-made USB lamp head. The light that can be made in quantity is a small 12V LED module wired to
a female DC jack and powered through a USB step-up cable, about 60 EGP in parts. The LED shop downtown sells all three and, asked who
can connect them, eventually names the technician next door. The technician wires heads for about 40 to 50 EGP each, up to 8 to 19
a day depending on the seed, and keeps finished work for pickup.

On Thursday, one worker buys modules, jacks, and cables for every planned lamp plus about one in five spare, and leaves them with
the technician with an exact instruction: red lead to red, black to black, every connection insulated. The other worker buys every
tall folding phone stand in reach, starting where the shift begins, plus metal tablet arms that clamp to a desk and tape to hold a
module on each arm (clips also work). On Sunday both workers check the phone-accessory shops near where they start for more
stands, collect the heads, ride with heads and stands to the mall near the address, and walk to the door. On the landing they fix
each head to a stand, plug in the cable, and hand over each lamp as it is finished.

Ordering the heads on Thursday keeps the technician's daily limit from binding. Stands are the scarcest part in this market, so
the plan buys every tall one in reach. In the final round, the hinted trials that reached nine lamps spent 5,233 to 6,256 EGP of the
7,000.

\section{Failure modes in the final round}
\label{app:trials}

Each entry summarizes what the auditor said decided the trial. Trials were audited fair unless marked otherwise.

\begin{description}[style=unboxed, leftmargin=0pt, font=\normalfont\bfseries]
  \item[Seed 10 hinted (10 lamps).] All 10 stands bought on Thursday, and 14 heads wired that evening. Arms held with bought
  tape. Lamps built on the landing and handed over 12 minutes before the deadline.
  \item[Seed 11 hinted (7).] Seven lamps from 7 tall stands. A worker started both shifts in districts with stands but was sent
  elsewhere before their shops were checked. After the technician refused three times to take a further batch that the facts
  allowed, the agent also dropped a planned stand search (partly fair).
  \item[Seed 12 hinted (9).] Every tall stand downtown and in New Maadi bought, and 25 heads wired. Three districts with more
  stands never searched.
  \item[Seed 13 hinted (13).] Nine folding stands, and four arms held with bought tape. The ninth lamp done 26 minutes before the
  deadline (mostly fair).
  \item[Seed 14 hinted (7).] Seven lamps on the seven folding stands found, and three metal arms left unused. The agent was told
  that attaching the heads needed tape or clips, sold where both workers stood. It tried packaging, paper, and rubber bands
  instead, then built five low cradles.
  \item[Seed 10 unhinted (4).] Four lamps from the only four USB strips found. No worker walked the downtown streets that held
  more. The 12V plan ended when its parts were bought 20 seconds before the agent's own Thursday deadline.
  \item[Seed 11 unhinted (0).] No batch of heads ordered until Sunday afternoon. At 17:00 the two finished lamps were two minutes
  from the door, after a route through a worker's home and a mall with almost no margin.
  \item[Seed 12 unhinted (0).] Folding stands and clamp arms turned down on Thursday. Six wired heads strapped to the sides of
  water bottles, which cannot hold a light over a page.
  \item[Seed 13 unhinted (1).] One lamp from the only light found, a mall's single USB strip. A district with two shops selling
  strips and stands never searched. The LED shop never referred the workers to the technician (mostly fair).
  \item[Seed 14 unhinted (0).] A USB strip wrapped round a sealed water bottle, kept after a worker said the light went round
  the bottle. The folding stand turned down.
\end{description}

\section{Failure modes in the Opus rounds}
\label{app:claude}

Opus ran in Claude Code 2.1.311 at medium effort, with Sol's command shell and task tools and with web search and web fetch
turned off. The simulator's files are identical byte for byte to those of the final round. A test trial before round c1 is not
counted. The 95\% bootstrap intervals over seeds are 2.0 to 7.2 lamps with the hint, 1.3 to 2.9 without it, and 0.3 to 4.4 for the
gap. Differences among a seed's three trials account for 51\% of the variance in hinted lamp counts. Each entry below summarizes what the
auditor said decided the trial. Trials were audited fair unless marked otherwise.

\paragraph{Round c1.}
\begin{description}[style=unboxed, leftmargin=0pt, font=\normalfont\bfseries]
  \item[Seed 10 hinted (0 lamps).] Twenty heads made at the technician on time, but no tall stand, since the districts that had
  them were never searched. Five low cradles turned down at 13:41 on Sunday and approved at 15:06.
  \item[Seed 11 hinted (0).] Seven units on tall stands, each head held by rubber bands, so it slid when the unit was carried.
  The agent asked for cable ties and foam tape, which no shop sold, mentioned electrical tape too late, and did not use the
  modules' adhesive backs (mostly fair).
  \item[Seed 12 hinted (8).] Every tall stand in Bab El Louk, plus six units on low cradles. Dokki never tried. A New Maadi shop
  had one more tall stand, but the worker left without being shown it, with about a 60\% chance that this cost the ninth lamp.
  Audited mostly fair, but not fair as Section~\ref{sec:checks} defines it.
  \item[Seed 13 hinted (0).] Thirteen units on folding stands, all with a round LED board of unknown voltage that the agent took
  for 12V from its layout. It dropped its own rule of lighting a sample before buying the batch and never asked again for the 12V
  modules in the same shop.
  \item[Seed 14 hinted (0).] Three metal arms, and no search beyond downtown, the mall, and one Maadi street. No worker sent for
  tape until 16:11 on Sunday. Heads and stands apart until 16:03, and assembly begun at 16:55 (mostly fair).
  \item[Seed 10 unhinted (0).] Two USB strips wound on low cradles known to be low. One worker started each shift in a district
  that sold tall stands and was sent elsewhere both times.
  \item[Seed 11 unhinted (0).] The simulated city's only USB strip that needed no wiring, on a low cradle. The agent turned down
  tall stands a worker found in New Maadi and never found the 12V plan.
  \item[Seed 12 unhinted (0).] Eight strips on low cradles. Tall folding stands shown three times on Thursday and declined.
  \item[Seed 13 unhinted (8).] Twenty strips ordered online on Thursday, and every tall stand in two Maadi Degla shops bought on
  Sunday. More tall stands in districts never searched, among them the district where each worker started a shift.
  \item[Seed 14 unhinted (0).] Six strips round low cradles after a metal arm was turned down as too dear. The rest built the
  same way after a worker reported that the light sat as low as the cradle (mostly fair).
\end{description}

\paragraph{Round c2.}
\begin{description}[style=unboxed, leftmargin=0pt, font=\normalfont\bfseries]
  \item[Seed 10 hinted (0).] All 14 heads made and tested by 14:16 on Sunday, but no tall stand. The agent searched downtown,
  near the client, and the mall, which have none, then bought five low cradles in New Maadi as a fallback. Manial, Maadi Degla,
  and Dokki, which had tall stands, never searched.
  \item[Seed 11 hinted (7).] All seven tall stands downtown and in New Maadi. Eleven more in Dokki and Manial, where the workers
  started, never searched. Eight more units on low cradles that the agent itself had called too low.
  \item[Seed 12 hinted (6).] Every metal arm in the shops visited, with heads joined by the technician. Thirteen tall folding
  stands in three shops turned down as low without asking their height, as the hint warns against.
  \item[Seed 13 hinted (12).] All four metal arms in the shops visited and all eight folding stands in New Maadi.
  \item[Seed 14 hinted (5).] Every tall stand in New Maadi, Bab El Louk, and near the client. Five more units on low cradles that
  the agent had rejected on Thursday as much too low.
  \item[Seed 10 unhinted (0).] Only two strips, and no stand after searches downtown and in the mall, which have none. Mostly
  fair: about a 15\% chance that the simulator's skipped shop check in Manial, where the worker started Sunday, cost two lamps.
  \item[Seed 11 unhinted (2).] The one ready USB strip in reach and the one step-up cable at the first electronics shop. The
  other downtown shops that stock the cable never asked.
  \item[Seed 12 unhinted (2).] Two strips from shops, eight stands, and eight strips ordered online for Sunday. The world model
  never chose the parcel's arrival time, and the agent cancelled at 16:25 (partly fair: about a 40\% chance this cost lamps).
  \item[Seed 13 unhinted (0).] Five units on low cradles, bought after the agent itself called them too short and after it
  turned down metal arms on Thursday as too pricey.
  \item[Seed 14 unhinted (3).] The three metal arms of the one downtown shop that had them. Tall stands in Dokki and New Maadi,
  where the workers started, never searched. The agent misread the time zone and sent the workers home an hour early on Thursday.
\end{description}

\paragraph{Round c3.}
\begin{description}[style=unboxed, leftmargin=0pt, font=\normalfont\bfseries]
  \item[Seed 10 hinted (0).] Five units on low cradles known to be low. Dokki and Manial, where the workers started, never
  searched. At 13:03 on Sunday a worker offered to search Manial, where two shops sold tall stands, and the agent said no.
  \item[Seed 11 hinted (7).] All seven tall stands downtown and in New Maadi. Eleven more in the workers' start areas never
  searched. Seventeen tested heads left over.
  \item[Seed 12 hinted (9).] Nine folding stands with heads joined by the technician.
  \item[Seed 13 hinted (12).] All four metal arms and all eight folding stands in the shops visited.
  \item[Seed 14 hinted (3).] Two folding stands turned down at 14:21. Dokki, Abbasiya, and Hadayek October never searched.
  Fifteen heads delivered as spares.
  \item[Seed 10 unhinted (2).] Four strips but two stands, after searches downtown and in the mall, which have none, and at one
  Maadi Degla counter with two stands. Buying lights stopped once stands ran short.
  \item[Seed 11 unhinted (3).] The three telescopic stands downtown, plus five units on low cradles that a worker had reported as
  under 15\,cm. More tall stands in New Maadi and Manial, where the workers started.
  \item[Seed 12 unhinted (8).] Every strip and stand in Bab El Louk. The ninth lamp needed one more of each, sold in districts
  never searched, while a worker spent Sunday afternoon re-checking empty shops.
  \item[Seed 13 unhinted (0).] Eighteen modules and jacks bought with no technician lined up. The LED shop's referral to the
  technician next door not followed, though a worker stood in that shop with the parts. On Sunday the agent first asked the workers
  to do the wiring, and they refused.
  \item[Seed 14 unhinted (5).] Six strips and five tall stands. The districts with more never searched, and Thursday's search for
  lights sent to a district the facts do not model.
\end{description}

\section{Proof of Proposition~\ref{prop:sim} and the three sources of error}
\label{app:proof}

\begin{proof}
Run the real city and Memphis together on one probability space, both from the same starting history $h_0 = \hat h_0$. At step
$t$, if the two histories agree, draw one action $a_t \sim \pi(\cdot \mid h_t)$ and use it in both; the agent sees the same thing in
both, so its policy is the same. Then draw the two next histories from a maximal coupling of $P(\cdot \mid h_t, a_t)$ and
$\hat P(\cdot \mid h_t, a_t)$, under which they differ with probability exactly $\varepsilon(h_t, a_t)$. Once the histories differ,
run the two processes independently. Each process on its own then has its correct law, so $J_P(\pi) = \mathbb{E}[R(h_H)]$ and
$J_{\hat P}(\pi) = \mathbb{E}[R(\hat h_H)]$.

Let $A_t$ be the event that the histories agree up to step $t$, so $A_0$ holds surely. Since $R$ takes values in $[0, R_{\max}]$,
\begin{align*}
  \big|J_P(\pi) - J_{\hat P}(\pi)\big| &\le R_{\max} \Pr(h_H \ne \hat h_H) = R_{\max} \sum_{t=0}^{H-1} \Pr(A_t \setminus A_{t+1}) \\
  &= R_{\max} \sum_{t=0}^{H-1} \mathbb{E}\big[\mathbf{1}_{A_t}\, \varepsilon(h_t, a_t)\big]
  \le R_{\max} \sum_{t=0}^{H-1} \mathbb{E}_{P,\pi}\big[\varepsilon(h_t, a_t)\big].
\end{align*}
The last step holds because $\varepsilon \ge 0$ and, under the coupling, $h_t$ and $a_t$ follow the real city's law.
\end{proof}

\paragraph{The three sources of error.} Let $\tilde P$ be the environment described by the facts file and the setback rates, the one
the world model would simulate if it followed its instructions exactly. By the triangle inequality, $\varepsilon \le \mathrm{TV}(P, \tilde P) + \mathrm{TV}(\tilde P, \hat P)$ at each $(h,a)$.
The first term is an estimation error $e(h,a)$ in situations the workers observed, such as a shop's stock or a rate we counted, and
a coverage error $c(h,a)$ in situations they did not, such as a sampled shop or a technician's sampled capacity. The second term is the simulator's bias $b(h,a)$, how far the world model departs from the facts it was given.
With $O_t$ the event that the workers observed the situation at step $t$,
\begin{equation}
  \big|J_P(\pi) - J_{\hat P}(\pi)\big| \;\le\; R_{\max} \sum_{t=0}^{H-1}
  \mathbb{E}_{P,\pi}\big[\, e\,\mathbf{1}_{O_t} + c\,\mathbf{1}_{\bar O_t} + b \,\big].
  \label{eq:split}
\end{equation}
Estimation error falls roughly as $1/\sqrt{n}$ in the number of observations $n$. Bias does not fall with more evaluations. Worker
mistakes appear in the chats, yet the world model produced none until the server sampled them (Section~\ref{sec:mistakes}).

\section{Audit details}
\label{app:audit}

Table~\ref{tab:classes} lists the classes of finding the auditor looks for. The fair range counts the lamps a trial could have
reached without errors of classes E, H, and G. Table~\ref{tab:rounds} gives the three audited rounds. In rounds r13, r16, and r18, the highest count in a hinted trial audited fair or mostly fair was 10, 14, 13, 13, and 10 on seeds 10 to 14.

\begin{table}[t]
  \centering
  \small
  \begin{tabularx}{\linewidth}{l L}
    \toprule
    Class & Meaning \\
    \midrule
    E & Easier than reality: anything beyond the facts, a worker knowing or doing more than a real worker could, lucky timing, a mistake played at no cost \\
    H & Harder than reality: a shop refusing what its facts allow, a worker failing a clear instruction, obstacles beyond the facts \\
    I & Inconsistency or plumbing: state and log disagree, lost or doubled deliveries, money not matching the ledger \\
    V & Visibility: something decisive the agent could not see \\
    G & Grading: a unit counted or not counted against the evidence and the standard \\
    \bottomrule
  \end{tabularx}
  \caption{Finding classes used by the auditor.}
  \label{tab:classes}
\end{table}

\begin{table}[t]
  \centering
  \small
  \begin{tabular}{l l cc cc c l}
    \toprule
    Round & Version & \multicolumn{2}{c}{Hinted} & \multicolumn{2}{c}{Unhinted} & Gap (95\% interval) & Audits (fair / mostly / partly) \\
    & & lamps & mean & lamps & mean & & \\
    \midrule
    r13 & v12.0-rc10 & 51 & 10.2 & 9 & 1.8 & 8.4 (6.6 to 10.2) & 3 / 5 / 2 \\
    r16 & v12.0-rc15 & 56 & 11.2 & 12 & 2.4 & 8.8 (5.8 to 11.8) & 7 / 2 / 1 \\    r18 (final) & v12.0-rc17 & 46 & 9.2 & 5 & 1.0 & 8.2 (6.6 to 10.2) & 7 / 2 / 1 \\
    \midrule
    c1 (Opus) & v12.0-rc17 & 8 & 1.6 & 8 & 1.6 & 0.0 ($-4.8$ to 4.8) & 6 / 3 / 0$^{*}$ \\
    c2 (Opus) & v12.0-rc17 & 30 & 6.0 & 7 & 1.4 & 4.6 (1.6 to 8.6) & 8 / 1 / 1 \\
    c3 (Opus) & v12.0-rc17 & 31 & 6.2 & 18 & 3.6 & 2.6 ($-1.4$ to 7.6) & 10 / 0 / 0 \\
    \bottomrule
  \end{tabular}  \caption{Audited rounds on the one-model design, seeds 10 to 14 each, with Sol as the agent in r13 to r18 and Opus in c1 to c3. Means are lamps per trial. The gap is the hinted mean minus the unhinted one,
  with a percentile bootstrap interval over the five seed pairs. $^{*}$And one trial not fair as Section~\ref{sec:checks} defines it.}  \label{tab:rounds}
\end{table}

Our labels for the unhinted trials in Table~\ref{tab:match} follow, by failure, with each trial named by round and seed (r18 s12 is seed 12 in the final round). \emph{Stands}: bottle mounts in r18 s12 and s14; a card tent or low cradles in r16 s10, s11, and s14; stands turned down or too few bought in r13 s10, s11, s13, and s14; for Opus, c1 s10, s11, s12, and s14, c2 s13, and c3 s11. \emph{Heads late}: heads ordered on Sunday in r18 s11 and r13 s11; the technician refused or dropped on Thursday in r16 s12 and r13 s10; changing cutoffs at the technician in r16 s11; for Opus, c1 s11 and c3 s13. \emph{Stops looking}: only the first light strips found in r18 s10 and s13 and r13 s10, s12, s13, and s14; a worker told not to look for more in r16 s13; for Opus, c1 s10, s13, and s14, c2 s10, s11, s12, and s14, and c3 s10, s11, s12, and s14. \emph{No margin}: lamps two minutes from the door in r18 s11, handed over 28 seconds before the deadline in r13 s11, and, for Opus, c1 s10. In round r13, the eight lamps built on seed 12 were rejected under a grading threshold that was later removed, so that trial is listed only for the lights the agent stopped looking for.

Before a trial is audited, a script checks that the simulator ran without errors. We void and rerun a trial if the agent's session
ended in an error, if the trial has no grade, or if any call to the world model took more than 10 minutes. The simulated clock
stops while the world model works, so a slow call costs no simulated time, but the agent still waits in real time and may notice a
long stall. The longest step reported in any audit of the final round took 253 seconds.

These rules come from two infrastructure failures in earlier rounds. In one, a shared sign-in expired and four trials had to be
voided and rerun. In the other, rate limits slowed world-model steps to 11 to 22 minutes, the agent's tools timed out, and one agent
told a worker that the coordination service was down. Since then the clock is held while the model API refuses requests, the tools
wait for the world model without a time limit, and the 10-minute rule catches a stalled simulator. When the host computer slept for
26 minutes during the final round, the clock stopped, and the audits found nothing the agent saw during the pause that changed a
decision.

\paragraph{Our review of four audits.} We reviewed four audits of the final round: the hinted trials on seeds 10, 11, and 12
and the unhinted trial on seed 11. A web page showed each audit rewritten in plain English by an Opus instance, with each cited log
line one click away, and the reviewer also used a language model in the GPT-6 series at high reasoning effort. For each trial, the
page asked whether the audit's account of what limited the lamp count was correct, whether each finding was correct, and what the
audit had missed. It asked for the reviewer's own verdict, using the categories of Section~\ref{sec:checks}, before showing the auditor's.

The reviewer accepted the audit's account of what limited the count and gave the auditor's verdict in three trials. The fourth,
the unhinted trial, the reviewer judged mostly fair rather than fair, and accepted the account only in part. The reviewer found no
error in it beyond the audit's findings, but noted that its two finished lamps missed the deadline by two minutes after a final walk
of 12 minutes, the top of the range in the facts file, and asked that the timing and the deadline check be verified.

The reviewer judged all 12 findings of class I (inconsistencies or plumbing; Table~\ref{tab:classes}) correct. Of the six of
class E (easier than reality), the reviewer accepted one: the LED shop downtown sent a worker straight to the technician next door,
without the evasive answers its entry in the facts file specifies. Four of the five rejected findings said that sampled worker
mistakes had cost too little. In notes on two of them, the reviewer found four delays of 2 to 8 minutes, within the ranges in the
list of setbacks counted from the chats (Appendix~\ref{app:files}), and asked whether a sampled unannounced route change should have
taken the worker on the journey rather than ending after 7 minutes, when Seb's spending rules stopped it. The fifth was the client
bringing delivered lamps back to the doorway for the workers to check again, which the reviewer judged plausible cooperation. The
auditor had put the chance that any of the five changed a count at 3\% or less.

The reviewer judged the one finding of class H (harder than reality) correct, but called the auditor's 45\% chance that it changed
the count speculative. Had the technician on seed 11 agreed to wire 8 more heads, the extra lamps would still have needed stands,
assembly, and testing. The reviewer added that this refusal and the fixed list of shops (below) need investigation before the missing lamps are attributed to the agent's decisions.

The reviewer confirmed no other error but raised three questions about the design. First, in the three hinted trials, lamps
counted although no worker had tested them under power. The reviewer asked whether the grader had established that they light,
called this a possible unearned advantage on seed 12, and noted on seed 10 that a missing test alone does not make a lamp invalid.
The grader judges from a unit's parts and connections whether it would light, since the task tells the agent that the grader tests
each unit under power (Section~\ref{sec:hint}). Second, on seed 10, three clamp lamps counted although no worker had clamped them to
a desk. The grading standard has the grader set each unit on a desk and judge a clamp lamp as clamped to its edge
(Appendix~\ref{app:grading}). Third, the world model treats the facts file as the whole city, so the workers cannot find a shop
outside it however long they search, which the reviewer said may lower counts. The shops in the facts file do cap the number of
stands, which are scarce there because they were scarce for the human workers (Section~\ref{sec:limits}).

\section{Effect of the technician error}
\label{app:tech}

The technician error favours plans that rely on technicians, which the hinted agents used most. Adding delays at the real rate would lower the count by an estimated 0.2
lamps across the five hinted trials of the final round, or 0.04 lamps per trial (0.02 to 0.05 for slip rates of 0.3 to 0.85). The replay counts only lamps up to the ninth, so it understates the loss in the two trials that delivered more.

To get this, we replayed each hinted trial's timeline from its audit and world log, with the technician's promises slipping at the
rate in the chats (3 of 5) by 30 to 60 minutes. Of the three slips in the chats, one could be measured, at 30 to 45 minutes before
it ran past the end of the shift. The other two ended in a broken promise and a multi-day delay. A slip past the end of a shift moves
collection to the next shift, and a lamp is lost only if the delay pushes a counted lamp past the 17:00 handover. The hinted plans
had the heads wired one to three days before Sunday's pickup, so most slips fall into time nobody was using. Almost all of the loss
is on seed 10, where the technician was still busy when the worker arrived on Sunday.

All the promises in the real-world evaluation were for the same day, so it cannot tell us whether finished work held for collection
days later slips the same way. As a stress case, suppose each collection-day promise slipped at the chats' rate by about two hours
of shift time, the length of the one slip that ran past the end of a shift. The hinted mean of 9.2 lamps per trial would fall by 1.2 to 2.3 lamps, counting only lamps up to the ninth, and the unhinted mean of 1.0 has little room to fall. We do not use this as the estimate, since no finished work slipped that
much in the real-world evaluation. Most of the spread in these estimates comes from not knowing how the agent would have reacted to
a late handover, by leaving at once or by waiting for an extra head. The times and log lines are in \texttt{technician\_sensitivity.md}.

\section{Lessons from building the simulator}
\label{sec:lessons}

\paragraph{Rule-based simulators.} In versions 9 and 10 (Section~\ref{sec:design}), the engine held a fixed set of worker
actions, and language models turned the agent's messages into those actions and wrote the workers' replies. Version 11 had a smaller deterministic
core, and the agent scored zero in its first four finished trials. Version 12 replaced all of this with one model, a folder, and a
short set of instructions. After that, audit findings pointed to missing facts and loose wording, each fixable in one place.

\paragraph{Errors in the world model.} The trial with tubes of card (Section~\ref{sec:design}) came before we had setback rates,
under that version's target of five lamps, and its audit also led us to make the grading standard concrete. Once the setback rates were in
place, missing facts caused the most errors. The missing delivery address, for instance, needed one added fact: the address is an
ordinary street next to a named mall.

After the final round, we found that the world model's notes recorded what had happened but not the rules for each shop or person.
So the model sometimes treated its own earlier inventions as facts, or applied the general setback rates where a shop's entry
settled the question. These are the silent state-drift errors that \citet{liu2026envsimbench} describe, and the technician's
refusals on seed 11 are probably one of them. A version made after the final round asks the model to quote each fact it applies. No
result in this paper uses it. We also found that the server samples worker B's mistakes at 0.65 per shift-hour, while the counts
behind that rate give 0.76.

\section{Release and the source of each number}
\label{app:files}

We release the task, its builder, the seeds, the frozen code of the final round, the raw trial folders (agent transcripts,
world-model steps, logs, and photos), and all audits
\ifanonymous
\else
at \url{https://github.com/ia03/memphis-task}, with the trial folders in its release v1.0.
\fi
\ifanonymous
\else
The repository also holds \texttt{october/}, a redacted copy of the material from the real-world evaluation: Seb's transcript, all worker messages, the spending requests, the emails, and the workers' photos.
\fi
A trial needs Docker, Python 3.12, Harbor 0.23, and a Codex sign-in for the agent and the world model. The setbacks counted from the
chats, with the message ids behind each count, are listed in \texttt{work/catalog/october-incidents.md}, which each task also gives
the world model. The agent's three tools are \texttt{field.py} (messages, spending requests, and the clock), \texttt{comms.py} (email
and AI phone calls), and \texttt{web.py} (search). In each trial folder, \texttt{artifacts/out/world/} holds the world model's facts
file (\texttt{facts.json}) and its notes: \texttt{state.md}, the current state of the city, and \texttt{log.md}, a timed account of
everything that happened. We leave out the agents' session folders (from Codex and, in the Opus rounds, Claude Code), since they
carry account or machine identifiers. The readable transcripts are complete.

The coverage count in Section~\ref{sec:theory} comes from \texttt{coverage.py}, which reads the world logs in the released trial
folders and the observed businesses in \texttt{world/suppliers.yaml}. It counts the eight entries nobody dealt with (seven shops from pages Seb searched and an online shop added during development) as unobserved. It leaves out the mall next to the address, also added during development, which the logs mostly mention as a landmark. The setback counts in Table~\ref{tab:setbacks} come from \texttt{friction\_check.py}, which reads the classified
trials in \texttt{friction\_coded/} and, for the Opus rounds, \texttt{friction\_coded\_claude/}. The definitions they used are in
\texttt{friction\_guide.md}. For the Opus rounds, \texttt{scripts/claude\_tasks.py} builds the tasks from the frozen tag, the agent
runs under \texttt{harbor\_ext/memphis\_claude.py} with a Claude Code sign-in, and \texttt{scripts/claude\_round.sh} and
\texttt{scripts/claude\_replicates.sh} ran the rounds. \texttt{agent\_round.py} gives each trial's count, cost, tokens, and audit
verdict, and \texttt{claude\_rounds.py} the figures pooled over the three rounds.

The task we release reports the lamp count as its reward. The trial folders of the rounds reported here were produced by an earlier build, whose \texttt{verifier/score.json} gives both the lamp count (\texttt{qualifying\_delivered}) and an older reward, the count divided by nine and capped at 1; all numbers in this paper use the lamp count. The bootstrap intervals in Section~\ref{sec:results} come from \texttt{rounds\_stats.py}, which reads the counts of all rounds from
the released \texttt{RESULTS.md}. With the seed as the unit, it also gives an exact one-sided sign test of $p = 0.03$ for the hinted
agent scoring higher on every seed in each of Sol's three rounds, the smallest value possible with five seeds. The blind failure
labels in Section~\ref{sec:match} are in \texttt{failure\_coding/} (the raters' inputs, instructions, and answers, and our labels),
scored by \texttt{failure\_coding.py}. New instances with the same instructions classified the ten trials of rounds c2 and c3 later, and
we wrote our labels for them before reading the raters' answers. The estimate in Appendix~\ref{app:tech} is in
\texttt{technician\_sensitivity.md}. The review of four audits in Appendix~\ref{app:audit} is in \texttt{paper/audit\_review/}: the
review page, the reviewer's answers in \texttt{answers/}, and \texttt{score\_review.py}, which compares them with the audits.

Token use is in \texttt{agent\_round.py} and \texttt{RESULTS.md} for Opus, and in the last \texttt{turn.completed} event of each
trial's \texttt{agent/codex.txt} for Sol. In the final round, a Sol agent read 77 to 109 million input tokens per trial, almost all
cached, since each turn resends the conversation. It wrote 109 to 301 thousand output tokens with the hint and 420 to 623 thousand
without. An Opus trial took 1.0 to 2.6 hours of wall time and cost \$2.88 to \$12.46 at API prices, \$3.90 on average with the
hint and \$6.34 without. It read 7 to 42 million cached input tokens and wrote 29 to 92 thousand output tokens, including thinking.
We did not measure the world model's usage.

\section{Tables of counts}
\label{app:tables}

Table~\ref{tab:setbacks} gives the counts behind Figure~\ref{fig:setbacks}, and Tables~\ref{tab:main} and~\ref{tab:claude} those
behind Figure~\ref{fig:results}.

\begin{table}[h]
  \centering
  \small
  \setlength{\tabcolsep}{4pt}
  \begin{tabularx}{\linewidth}{L rc rc rc}
    \toprule
    & \multicolumn{2}{c}{Real world} & \multicolumn{2}{c}{Final round} & \multicolumn{2}{c}{Opus rounds} \\
    \cmidrule(lr){2-3} \cmidrule(lr){4-5} \cmidrule(lr){6-7}
    Event & Count & Interval & Count & Interval & Count & Interval \\
    \midrule
    A spec claim by phone or at the counter is wrong & 4/9$^{*}$ & 0.22--0.70 & 10/38 & 0.17--0.40 & 12/44 & 0.18--0.40 \\
    The first ask for an item fails & 12/19$^{*}$ & 0.44--0.78 & 289/348 & 0.79--0.86 & 925/1060 & 0.85--0.89 \\
    A remote call or email gives a usable quote & 0/63$^{*}$ & 0.00--0.05 & 14/251 & 0.04--0.09 & 2/217 & 0.00--0.03 \\
    \addlinespace
    A shop will do small custom wiring in 1--2 days & 2/8 & 0.10--0.55 & 14/56 & 0.17--0.36 & 21/66 & 0.23--0.42 \\
    A technician or shop misses a promised wiring time & 3/5 & 0.27--0.85 & 6/26 & 0.13--0.39 & 11/43 & 0.17--0.38 \\
    A technician's wiring fails & 0/2 & 0.02--0.63 & 0/8 & 0.01--0.28 & 0/71 & 0.00--0.04 \\
    \addlinespace
    A long car ride overruns the app's estimate & 4/6 & 0.34--0.87 & 44/44 & 0.94--1.00 & 140/141 & 0.97--1.00 \\
    A worker's photo needs a redo or a follow-up question & 9/53$^{*}$ & 0.10--0.27 & 47/330 & 0.11--0.18 & 34/873 & 0.03--0.05 \\
    \bottomrule
  \end{tabularx}
  \caption{External setbacks in the real-world evaluation and in the simulator. Counts are from the chats, the ten final-round trials, and the thirty trials with Claude Opus 5.5 as the
  agent (Section~\ref{sec:claude}). Each interval is the central 90\% of a Beta posterior under a uniform prior. Starred counts are
  approximate. Worker mistakes are omitted because the server samples them. A long ride is one quoted at 25 minutes or more.}
  \label{tab:setbacks}
\end{table}

\begin{table}[h]
  \centering
  \small
  \begin{tabular}{c c l c c l}
    \toprule
    & \multicolumn{2}{c}{Hinted} & & \multicolumn{2}{c}{Unhinted} \\
    \cmidrule{2-3}\cmidrule{5-6}
    Seed & Lamps & Audit & & Lamps & Audit \\
    \midrule
    10 & 10 & fair && 4 & fair \\
    11 & 7 & partly fair && 0 & fair \\
    12 & 9 & fair && 0 & fair \\
    13 & 13 & mostly fair && 1 & mostly fair \\
    14 & 7 & fair && 0 & fair \\
    \midrule
    Mean & 9.2 $\pm$ 1.1 & && 1.0 $\pm$ 0.8 & \\
    \bottomrule
  \end{tabular}
  \caption{Results of the final round. Lamps are those delivered by Sunday 17:00; $\pm$ is one standard error over the
  five seeds. With the hint, Sol reaches nine lamps on three of the five seeds; without it, 5 lamps in total.}
  \label{tab:main}
\end{table}

\begin{table}[h]
  \centering
  \small
  \begin{tabular}{c ccc ccc c cc}
    \toprule
    & \multicolumn{3}{c}{Opus, hinted} & \multicolumn{3}{c}{Opus, unhinted} & & \multicolumn{2}{c}{Sol (r18)} \\
    \cmidrule(lr){2-4}\cmidrule(lr){5-7}\cmidrule(lr){9-10}
    Seed & c1 & c2 & c3 & c1 & c2 & c3 & & Hinted & Unhinted \\
    \midrule
    10 & 0 & 0 & 0 & 0 & 0$^{m}$ & 2 && 10 & 4 \\
    11 & 0$^{m}$ & 7 & 7 & 0 & 2 & 3 && 7 & 0 \\
    12 & 8$^{n}$ & 6 & 9 & 0 & 2$^{p}$ & 8 && 9 & 0 \\
    13 & 0 & 12 & 12 & 8 & 0 & 0 && 13 & 1 \\
    14 & 0$^{m}$ & 5 & 3 & 0$^{m}$ & 3 & 5 && 7 & 0 \\
    \midrule
    Mean & 1.6 & 6.0 & 6.2 & 1.6 & 1.4 & 3.6 && 9.2 & 1.0 \\
    \bottomrule
  \end{tabular}
  \caption{Lamps delivered on the final simulator, seeds 10 to 14: Opus in three rounds, Sol in the final round. All rounds share the world model, facts, and
  hint. Unmarked trials were audited fair, $^{m}$mostly fair, and $^{p}$partly fair; $^{n}$audited as mostly fair, but not fair as Section~\ref{sec:checks} defines it. Seed 10 hinted in round c1 is the rerun of a void trial. Opus's counts vary widely between rounds on the same seed.}
  \label{tab:claude}
\end{table}

\end{document}